\pdfoutput=1

\documentclass[11pt]{article}
\usepackage[final]{acl}

\usepackage{times}
\usepackage{latexsym}

\usepackage[T1]{fontenc}
\usepackage[utf8]{inputenc}

\usepackage{microtype}

\usepackage{inconsolata}
\usepackage{tocloft}
\usepackage{etoc}

\usepackage{graphicx}
\usepackage[dvipsnames]{xcolor}
\definecolor{pink1}{RGB}{230, 100, 150}
\definecolor{blue1}{RGB}{80, 140, 230}
\definecolor{orange1}{RGB}{255, 140, 80}
\definecolor{green1}{rgb}{0,0.8,0}
\usepackage{multirow}
\usepackage{colortbl,booktabs}
\usepackage{makecell}
\usepackage{tikz}
\usepackage{subcaption}
\usepackage{amsmath}
\usepackage{amsthm}
\usepackage{amssymb}

\usepackage{enumitem}
\usepackage{booktabs}
\usepackage{tabularx}
\usepackage[table]{xcolor}
\usepackage{colortbl}
\usepackage{wrapfig}
\usepackage{booktabs}
\usepackage{arydshln}
\usepackage{makecell}
\definecolor{pastelpurple}{RGB}{202,191,255}
\definecolor{pastelorange}{RGB}{255,223,186}
\usepackage[ruled,vlined]{algorithm2e}

\usepackage{xspace}
\usepackage{tcolorbox}
\usepackage{float}

\newcommand{\model}{TaSe\xspace}
\newcommand{\disenmodule}{TriDe\xspace}
\newcommand{\disenloss}{TriDe\xspace}
\newcommand{\desco}{DesCo\xspace}

\title{Talk in Pieces, See in Whole: Disentangled and Hierarchical Representation Learning in Language-based Object Detection}

\newcommand{\equalcontrib}{\textsuperscript{*}}
\newcommand{\corrauthor}{\textsuperscript{\textdagger}}

\author{
    \textbf{Sojung An\textsuperscript{1}\equalcontrib},
    \textbf{Kwanyong Park\textsuperscript{2}\equalcontrib},
    \textbf{Yong Jae Lee\textsuperscript{3}},
    \textbf{Donghyun Kim\textsuperscript{1}\corrauthor}
    \\
    \textsuperscript{1}Korea University\quad
    \textsuperscript{2}University of Seoul\quad
    \textsuperscript{3}University of Wisconsin--Madison\\
    \normalsize{
        \equalcontrib Equal contribution
        \qquad
        \corrauthor Corresponding author:
        \href{mailto:d_kim@korea.ac.kr}{d\_kim@korea.ac.kr}
    }\\
    \includegraphics[height=0.6em]{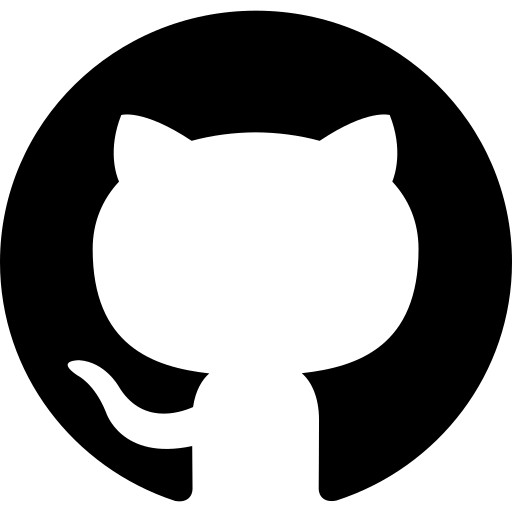}
    \normalsize\url{https://github.com/ssojungan/tase}
}

\begin{document}
\maketitle

\begin{abstract}
Vision-language models (VLMs) have advanced multimodal perception, demonstrated by open-vocabulary object detection with simple language queries. State-of-the-art VLMs still struggle to handle complex queries involving descriptive attributes and relational clauses.
To address this problem, we propose restructuring linguistic representations according to the hierarchical relations within sentences for language-based object detection. 
A key insight is that textual tokens should be disentangled into core components—objects, attributes, and relations—and aggregated into hierarchically structured sentence-level representations. Building on this principle, we introduce the \model (\textbf{Ta}lk in Pieces, \textbf{Se}e in Whole) framework with three main contributions: (1) a hierarchical synthetic captioning dataset spanning three tiers from category names to descriptive sentences; (2) the three-component disentanglement module guided by a novel disentanglement loss function, transforms text embeddings into subspace compositions; and (3) aggregating disentangled components into hierarchically structured embeddings guided by the proposed hierarchical objectives. 
Experimental results under the OmniLabel benchmark show a 24\% performance improvement, demonstrating the importance of linguistic compositionality.

\end{abstract}

\section{Introduction}
\label{sec:intro}

\begin{figure*}[t!]
    \centering
    \begin{minipage}{0.51\linewidth}
        \centering
        \begin{subfigure}[t]{\linewidth}
        \centering
        \includegraphics[width=\linewidth]{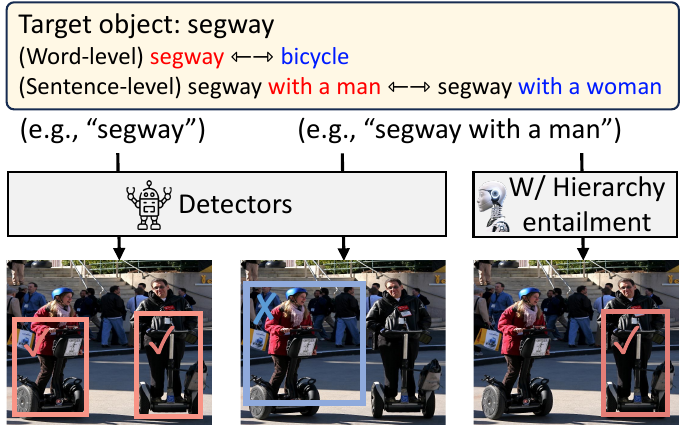}
        \caption{Detection results from w/o and w/ our hierarchical entailment}
        \label{fig:fig1_1}
        \end{subfigure}
    \end{minipage}%
    \hspace{3mm}
    %\hfill
    \begin{minipage}{0.42\linewidth}
        \begin{subfigure}[t]{\linewidth}
        \begin{tikzpicture}
        \node[inner sep=0pt, anchor=south west] (img) at (0,0) {
            \includegraphics[width=\linewidth]{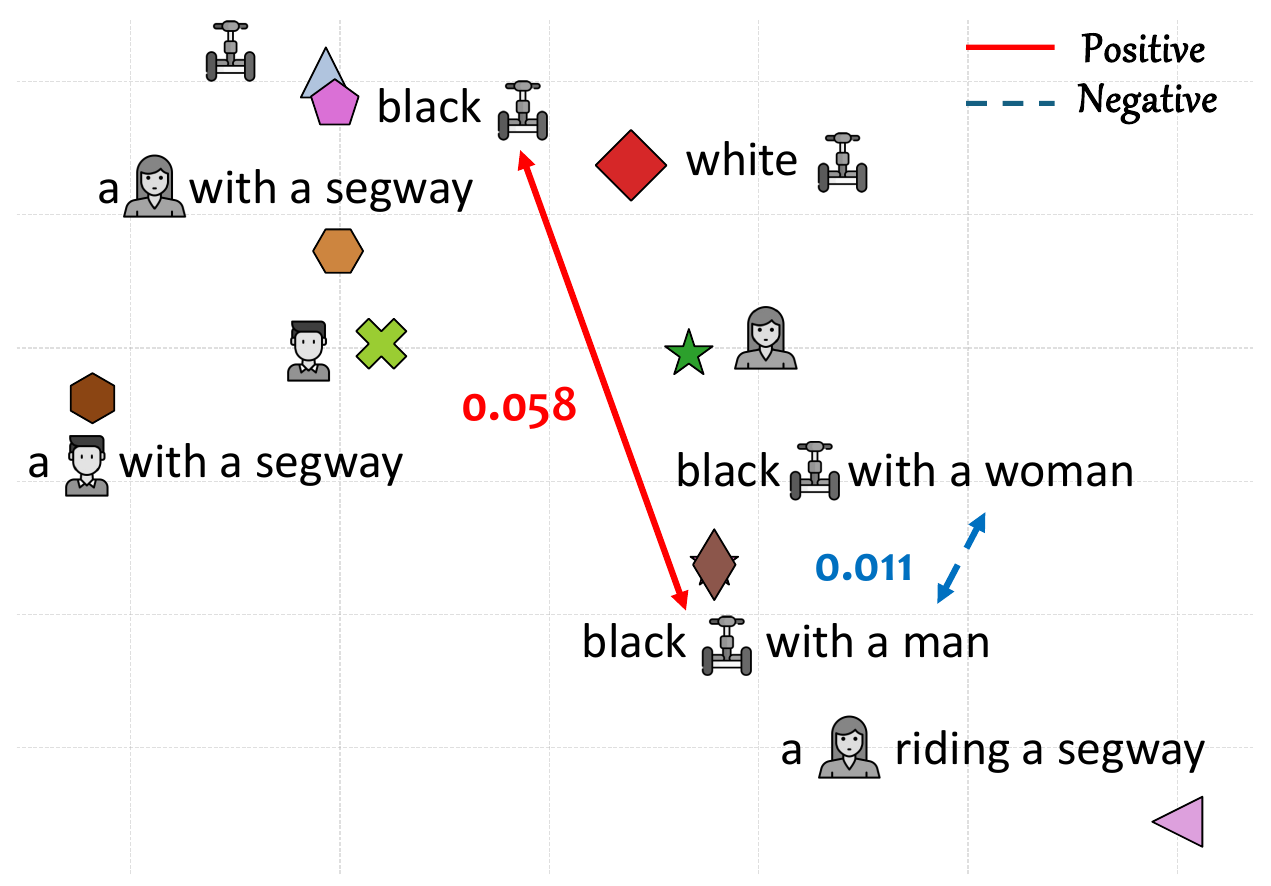}
        };
        \draw[black, thick] (img.south west) rectangle (img.north east);
        \end{tikzpicture}
        \caption{t-SNE of sentence features from the text encoder of VLMs}
        \label{fig:fig1_2}
        \end{subfigure}
    \end{minipage}
    \vspace{-2mm}
    \caption{(a) VL detectors struggle with attributes or relations due to limitations in capturing fine-grained semantics from image-text similarity. (b) Limitations of text encoders in VLMs for compositional understanding.}
    \label{fig:motivation}
    \vspace{-2mm}
\end{figure*}

Vision-language (VL) understanding is a long-standing and fundamental problem that requires learning modality-specific representations and cross-modal alignment.
Foundational VLMs such as CLIP~\citep{clip} have leveraged web-scale image-text pairs to learn generic VL representations, achieving strong generalization performance on tasks such as image classification and image-text retrieval.
Building upon these advances, recent studies have actively explored grounding language queries into specific image regions (e.g., open-vocabulary object detection~\citep{gdino, gen, rodmllm}. Many approaches~\citep{gdino, glip} distill the general VL knowledge embedded in foundational models into object detectors and have demonstrated remarkable results in detecting previously unseen object categories—commonly referred to as open-vocabulary object detection~\citep{gu2022openvocabulary}.
Current VL detectors succeed when input queries are short and consist of simple category names, while they struggle to comprehend complex language queries and accurately localize the corresponding objects. We further analyze this behavior using GLEE~\citep{glee}, a state-of-the-art visual grounding foundation model (see Fig.~\ref{fig:fig1_1}). The model reliably detects objects given simple noun phrases (e.g., ``segway''). VLMs fail when faced with more complex and specific queries (e.g., ``segway with a man'').
%indicating their limited compositional understanding.

To investigate the underlying cause of this limitation, we visualize sentence-level text features using t-SNE~\citep{van2008visualizing} in Fig.~\ref{fig:fig1_2}. Interestingly, we observe that although some sentences (``a segway with a man'' vs. ``a segway with a woman'') refer to different target objects (``with a man'' vs. ``with a woman''), their embeddings remain close in the feature space due to shared tokens (``segway'').
This contrasts with the ideal cases, where embeddings of distinct target objects should be well-separated, while those of the same object with different attributes should be closer for object detection (``a segway'' vs. ``a black segway'').
Such a mismatch is attributed to VL detector objectives~\citep{gdino, glip}, which rely on augmented negative caption supervision and encourage coarse text-region discrimination.
Augmented caption supervision~\citep{desco,yuksekgonul2023when} is effective for coarse-grained concept analysis but provides limited guidance for modeling fine-grained ambiguities among concepts.
Addressing these ambiguities requires sentence embeddings that capture compositional semantics by disentangling text into meaningful components.

In this paper, we propose a novel framework that disentangles component-wise text features (``Talk in Pieces'') and explicitly learns hierarchical knowledge (``See in Whole'') from these disentangled representations to construct sentence-level understanding for language-based object detection. 
We refer to our framework as \textbf{\model} (\textbf{Ta}lk in Pieces, \textbf{Se}e in Whole).
We begin by formally defining a hierarchical structure tailored for language-based object detection.
Specifically, we design our new HiVG dataset, a three-tiered hierarchy: object–attribute–relation, where the first tier represents object category names (i.e., segway), the second tier adds descriptive attributes (i.e., black), and the last tier includes relational phrases that describe interactions or contexts (i.e., with a woman).
We also construct negative samples following the same three-tier structure. 
This mitigates bag-of-words behavior driven by lexical overlap and supports reliable semantic discrimination between positive and negative prompts.
Our approach builds on phrase grounding datasets like Visual Genome~\citep{krishna2017visual}, which provide densely annotated phrases associated with images and object regions. Using a large language model (LLM)~\citep{dubey2024llama,yang2025qwen3,achiam2023gpt4}, we abstract these phrases into a three-tier hierarchy—object, attribute, relation—by sequentially removing relational and attribute information in phrases to obtain the final object categories.
 
To learn compositional sentence representations from the HiVG dataset, we disentangle it into the three components (``Talk in Pieces''): object, attribute, and relation.
The disentangled design maps text representations into semantic subspaces and captures hierarchical structure through separated semantic units.
We further design a lightweight learnable attention module for the \disenloss (Three-component disentanglement).
The key idea of \disenloss is to leverage the hierarchical structure of the HiVG dataset to contrast component-wise tokens, allowing targeted tokens to be adjusted without loss of meaningful information.
Then, we guide the model to learn linguistic representations that capture these levels of abstraction.
This facilitates learning of sentence context enriched with descriptive attributes and relational clauses.
We introduce a hierarchical aggregation method (``See in Whole'') based on sentence-level hierarchical entailment.

To summarize, our main contributions are as follows: 
1) We present an efficient hierarchical data generation pipeline that abstracts dense existing phrases into an explicit hierarchical structure of object–attribute–relation. 2) We introduce a framework for disentangling the three core components and employ the \disenloss loss to guide this process. 3) We propose a method for learning disentangled and hierarchical representations that capture sentence-level inductive biases and generalize across VL detectors. With hierarchical learning on our generated dataset HiVG, our model significantly outperforms strong baselines, including state-of-the-art VL detectors, on challenging language-based object detection benchmarks such as OmniLabel~\citep{omnilabel} and D3~\citep{d3}.

%-------------------------------------------------------------------------

\section{Related Works}
\label{sec:related_works}
\subsection{Language-based Object Detection}
\label{subsec:related_work1}
Language-based object detection aims to locate and identify objects in images using free-form text. 
One of the leading approaches is to transfer the pre-trained model and align images and texts using contrastive learning \citep{gao2024clip,glip,desco,park2024weak}. 
Contrastive learning enhances compositionality in VLMs by capturing relationships with contextual entities and improves the understanding of object relationships \citep{desco,owl,gu2022openvocabulary,gdino}. 
GLIP \citep{glip} proposes to add deep fusion layers between different modalities and learn a language-aware visual representation based on reformulated alignment scores. 

Existing approaches overlook the need for contextualized sentence-level understanding of VL text embedding.
For example, APE \citep{ape}; GLEE \citep{glee}; DINO-X \citep{dinox}; and \citet{zeng2024investigating} explore VLM alignment challenges and highlight the need to improve reasoning capabilities in multimodal large language models (MLLMs).
These works investigate model capabilities from restricted VL perspectives, with a primary focus on fine-grained textual details and inter-object relationships.
VL detectors still struggle to align images with syntactically intricate language queries \citep{wang2023equivariant}, underscoring the need for a more grounded contextual understanding of text.
While several works have explored disentanglement in visual components~\citep{bengio2013representation,wang2024disentangled,qi2024deadiff}, disentangled and hierarchical textual representations remain largely underexplored for VL models, particularly in the context of language-based object detection.

%-------------------------------------------------------------------------

\subsection{Hierarchical Entailment for Vision-Language Models}
\label{subsec:related_work2}
To better represent the embedding space of VLMs, hyperbolic learning has highlighted the need to capture hierarchical structures and relationships in multimodal data.
Hyperbolic learning was formulated on the Poincar\'e ball by \citet{ganea2018hyperbolic}, learning entailment relations between embedded objects.
The formulation adopts the Lorentz model of \citet{lou2020differentiating} to avoid computationally expensive gyrovector operations.
Hyperbolic learning maps the embedding into an entailment cone (EC) to represent hierarchical entailment in a continuous space.
Recent studies investigated the use of the EC embedding for vision tasks \citep{atigh2022hyperbolic,kong2024hyperbolic,khrulkov2020hyperbolic}, multimodal learning \cite{desai2023hyperbolic,hong2023hyperbolic,pal2025compositional}, and synthetic data generation \citep{kong2024hyperbolic}.
However, the hyperbolic manifold needs to transpose features from Euclidean to hyperbolic and requires additional hyperparameter configurations.
\citet{alper2024emergent} propose radial embedding (RE) optimization for learning hierarchical representations directly in Euclidean space. 
Inspired by this approach, we extend RE optimization to language-based object detection via sentence-level hierarchical representation learning.

%-------------------------------------------------------------------------

\begin{figure*}[t]
   \centering
   \includegraphics[width=\linewidth]{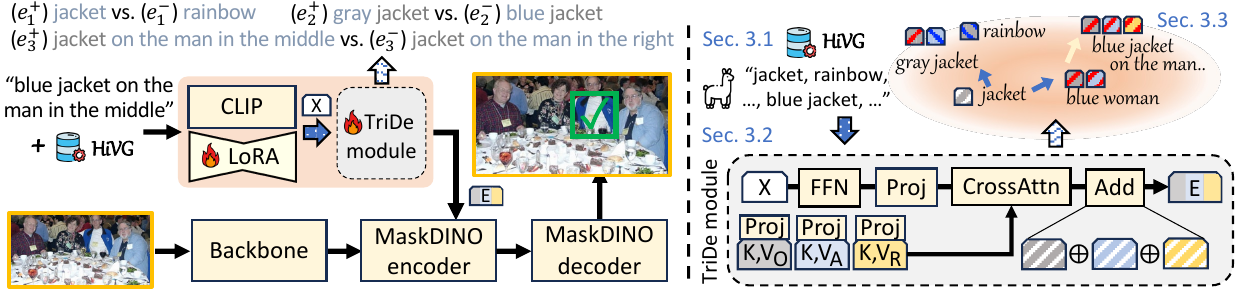}
   \vspace{-5mm}
   \caption{The overall framework of \model. (Left) The text encoder is fine-tuned with LoRA~\citep{hu2021lora} and the \disenmodule module to restructure text representations. (Right) \textbf{Top:} Hierarchy aggregated embeddings with HiVG, where the re-captioned dataset passes through the \disenmodule module to learn linguistic hierarchy. \textbf{Bottom:} Architecture of the \disenmodule module.}
  \label{fig:method}
  \vspace{-2mm}
\end{figure*}

\section{TaSe: Disentangled and Hierarchical Text Representation Learning for Language-based Object Detection}
\label{sec:methodology}
\vspace{-2mm}
This section introduces \model (\textbf{Ta}lk in Pieces, \textbf{Se}e in Whole), a framework for disentangling and hierarchical aggregation.
Specifically, our approach comprises three components: 1) the HiVG dataset (Sec.~\ref{subsec:re_captioning}), a synthetic dataset re-captioned from VG to capture hierarchical entailment relations; 2) disentangling text representations into object, attribute, and relation subspaces to align semantic pieces within sentences (Sec.~\ref{subsec:drl}); and 3) a hierarchical aggregation method to represent contextualized sentence embedding based on disentangled tokens (Sec.~\ref{subsec:text_hierarchy}).
Fig.~\ref{fig:method} outlines how \model learns contextualized sentence representations for language-based object detection.
\subsection{HiVG: Hierarchy Captioning Pipeline}
\label{subsec:re_captioning}

Open-vocabulary detectors often rely on keywords and fail to distinguish target objects from their attributes and relations because they lack hard textual negatives that reflect linguistic hierarchy. To address this problem, we propose a \textbf{Hi}erarchical captioning pipeline that re-captions the \textbf{V}isual \textbf{G}enome dataset (\textbf{HiVG}) by leveraging LLMs (i.e., Llama-3-8B \citep{dubey2024llama}, Qwen3-8B \citep{yang2025qwen3}, and GPT-4-nano \citep{achiam2023gpt4}) and lexical databases (e.g., WordNet \citep{miller1995wordnet} and ConceptNet \citep{speer2017conceptnet}). 
HiVG is a synthetic dataset constructed by spanning from category names to descriptive sentences and structuring hierarchical captions into three tiers: objects, attributes, and relations. 
Each caption in the Visual Genome~\citep{krishna2017visual} annotation is transformed into three positive ($e^{+}$) and negative tiers ($e^{-}$) where $e$ follows the notation introduced in Sec.~\ref{subsec:text_hierarchy}. 
\begin{enumerate}[label=\textbullet, itemsep=0pt, parsep=0pt, topsep=0pt, partopsep=0pt, leftmargin=1em]
    \item Tier 1. Category names (object): containing the class name (e.g., \textbf{jacket ($e_1^{+}$)} and \textbf{rainbow} ($e_1^{-}$)).
    \item Tier 2. Enriched descriptions (w/ attribute): adding an attribute to the object (e.g., \textbf{blue} ($e_2^{+}$) and \textbf{red} ($e_2^{-}$)) for learning fine-grained linguistic compositionality.
    \item Tier 3. Contextual understanding (w/ attribute and relation): emphasizing the relationships between objects by injecting a relation into the second-tier caption (e.g., \textbf{on the man in the middle ($e_3^{+}$)} and \textbf{on the man in the right} ($e_3^{-}$)).
\end{enumerate}
The core idea disentangles the category names from descriptive components (e.g., attributes and relations) in the text embedding space.
The model focuses on the detection target rather than descriptive components to mitigate false positives.
Further details are provided in Appendix~\ref{sec:data_details}.

\subsection{Talk in Pieces: Component-Wise Text Disentanglement}
\label{subsec:drl}
Textual descriptions typically contain not only descriptive attributes but also complex relational structures, which cause false positives in language-based object detection.
To address this, we propose the \disenmodule module to disentangle text embeddings into meaningful subspaces, which adaptively refines these components to enhance semantic representation.

\begin{figure*}[t]
    \centering
    \includegraphics[width=\linewidth]{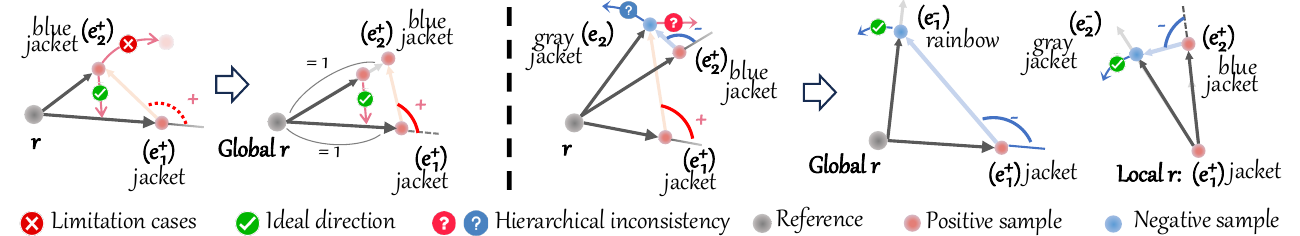}
    \caption{Limitations of RE and our solutions. (Left) Geometry-preserving entailment. Variations in vector magnitude can distort angular measurements, leading to inconsistent hierarchies. Normalization ensures the hierarchy depends on directional relationships. (Right) Local-reference entailment contrast. Under the global reference, higher tiers contain positive (``blue jacket'') and negative (``gray jacket'') samples that share semantic components (``jacket''), which leads to hierarchical inconsistency; a local reference better separates embeddings.
    }
    \label{fig:loss_problem}
\end{figure*}

\noindent\textbf{Text embedding.}
We extract text features using a CLIP text encoder with low-rank adaptation (LoRA) \citep{hu2021lora} for efficiently evolving text embeddings from the text encoder. 
Let $\{v_i, t_i\}_{i=1}^{B}$ be a batch of image–text pairs. 
The text embedding $\mathbf{X} = \mathcal{T}_{\theta}(t_i)$, where $\mathrm{X} \in \mathbb{R}^{B \times T \times d_{\text{model}}}$, is obtained by the text encoder of CLIP.
Here, $B$, $T$, and $d_{model}$ denote the batch size, number of tokens, and embedding dimension, respectively.
A text projection layer maps the input into $\mathrm{X} \in \mathbb{R}^{B \times T \times D}$, where $D$ denotes the text embedding dimension.

\noindent\textbf{Component-wise disentanglement.}
We disentangle text representations into three components—objects, attributes, and relations. 
This design mirrors the three-tier structure of HiVG and facilitates the learning of effective contextualized sentence embeddings.
We adjust learnable vectors $\mathrm{V_{O}}, \mathrm{V_{A}}, \mathrm{V_{R}} \in \mathbb{R}^{B\times T\times D}$ to disentangle the text embedding into the three components.
We employ a multi-head cross-attention layer between the learnable vectors and text embedding $\mathrm{X}$.
Let FFN, LN, and Proj denote the feed-forward network, layer normalization, and projection layer, respectively.
The \disenmodule module is defined as follows:
\begin{equation}
    \begin{array}{c}
    \mathbf{X} = \text{LN}(\text{Proj}(\mathrm{X} + \text{FFN}(\mathrm{X}))),\\
    \left[\text{O}, \text{A}, \text{R}\right] = \text{CrossAttn}(\mathrm{X}, \left[\mathrm{V}_{\text{O}}, \mathrm{V}_{\text{A}}, \mathrm{V}_{\text{R}}\right]),\\
    \mathbf{E} = \text{pooling}(\text{FFN}(\text{LN}(\mathrm{O} + \mathrm{A} + \mathrm{R}))),\\
    \end{array}
\end{equation}
where $\mathrm{O}$, $\mathrm{A}$, $\mathrm{R}$, and $\mathrm{E}$ represent the object, attribute, relation components, and restructured text embedding.
Note that $\mathrm{E}$ is employed to learn hierarchical entailment for contextualized sentence representations in Sec.~\ref {subsec:text_hierarchy}.
The CrossAttn($\mathrm{Q}, \mathrm{V}$) with the scaling factor $d_{k}$ is defined as follows:
\begin{equation}
\resizebox{0.88\columnwidth}{!}{$
    \begin{array}{c}
    \text{CrossAttn}(\mathrm{Q}, \mathrm{V})=\text{Softmax}\Big(\dfrac{\text{Proj}(\mathrm{Q})\text{Proj}(\mathrm{V})^{T}}{\sqrt{d_{k}}}\Big)\text{Proj}(\mathrm{V}).
    \end{array}$}
\end{equation}
Here, $\mathrm{Q}$ denotes the \textit{query}. The \textit{key} and \textit{value} are identical in this formulation, so only the $\mathrm{V}$ term is used. $\mathrm{E}$ is mean-pooled to encode the sentence-level representation used for alignment with visual embeddings.
The detailed algorithm is described in Appendix Alg.~\ref{alg:compositional_learning}.

\noindent\textbf{Objective for component disentanglement.}
We found that directly increasing the angular distance for negative captions amplifies the effect of shared lexical content, leading to unstable gradients and sharp loss escalation.
The disentangle–aggregate design separates semantic components and recombines them, and yields stable optimization of angular relationships in the embedding space.
Components are aligned with their positive counterparts, while negatives are enforced to remain distant according to their tier.
Let $t$ be the tier of HiVG, and the disentanglement of text embedding is adjusted $\mathcal{L}_{\text{\disenloss}}$ as follows:
\begin{equation}
%\small
\resizebox{0.85\columnwidth}{!}{$
\begin{array}{c}
\mathcal{L}_{\text{\disenloss}} =
\lambda \!\sum_{(i,j)\in\{(O,A),(O,R),(A,R)\}}\! |\mathbf{i}\!\cdot\!\mathbf{j}| \\[4pt]

+ \sum_{X\in\{O,A\}}\sum_{t\in T_X}
\Big(m + \cos(X_t^+,X_{t+1}^+) \\
- \cos(X_t^+,X_t^-)\Big) + \Big(m-\cos(R_l^+,R_l^-)\Big).
\end{array}$}
\end{equation}
$\mathcal{L}_{\text{\disenloss}}$ and $m$ represent compositional loss for the \disenmodule module and margin, respectively.
$\lambda$ represents a hyperparameter for stability adjustment.
Minimizing correlation among components promotes an inductive bias toward semantically grounded object representations.

\subsection{See in Whole: Sentence-Level Hierarchical Aggregation}
\label{subsec:text_hierarchy}
We design a hierarchical aggregation method for disentangled features to serve fine-grained semantic distinction, which helps capture contextual meaning beyond simple word-level perception.

\subsubsection{Background: hierarchical entailment in Euclidean space}
The goal of hierarchical entailment is to learn general concepts by representing entailment relations via low-dimensional embeddings \citep{ganea2018hyperbolic}.
Conventional contrastive learning models learn relations between positive and negative samples, whereas the radial embedding (RE) objective~\citep{alper2024emergent} encodes hierarchical structure through exterior angles with respect to a reference point.
Two key advantages of this representation learning with the RE are: 1) modeling sentence-level structure; and 2) learning compositional generalization without requiring transformation into sphere space.
Let $\Xi$ denote the exterior angle in radians and $r\in\mathbb{R}^d$ denote a root embedding, the exterior angle between embedding $a$ and $b$ in the RE objective is defined:
\begin{equation}
    \begin{array}{c}
    \Xi\langle a, b\rangle = \cos^{-1} \left( \frac{\mathbf{a'} \cdot \mathbf{b'}}{\|\mathbf{a'}\| \|\mathbf{b'}\|} \right) \leq \pi.\\
    \text{where} \ \mathbf{a'} = \mathbf{a} - r, \quad 
    \mathbf{b'} = \mathbf{b} - r - \mathbf{a'}. 
    \label{eq:eq1}
    \end{array}
\end{equation}
The value $\langle a, b\rangle_{\Xi} \in [0, \pi]$ is bounded.
Given a root embedding $r$ and a reference embedding of a text embedding $e$, the objective function of the RE is represented as follows:
\begin{equation}
\mathcal{L}_{RE} = \sum_{i} \left( \langle e_{i}^+, e_{>i}^+ \rangle_{\Xi} 
- \langle e_{i}^+, e_{i}^- \rangle_{\Xi} \right).
\vspace{-10pt}
\end{equation}
Here, $e_i^+$ and $e_{>i}^+$ denote distinct positive embeddings, and $e_i^-$ is a corresponding negative. 
The objective $\mathcal{L}_{RE}$ encourages smaller exterior angles between positive pairs while enforcing larger angles between positive and negative pairs.
This deviation reflects a misalignment from the reference anchor $r$ and corresponds to a larger angular distance in the embedding space.

\subsubsection{Local-reference hierarchical entailment}
The existing exterior-angle formulation presents two limitations for language-based object detection: \textbf{Radial distance} from the root alters the exterior angle and distorts the embedding geometry; \textbf{Fixed-root hierarchical learning} causes conflicts between positive and negative embedding regions.
The proposed hierarchical entailment objective addresses these limitations with two components: 1) geometry-preserving entailment and 2) local-reference entailment contrast.\\
\textbf{Geometry-preserving entailment (Fig.~\ref{fig:loss_problem} Left)}. 
Sentence-level hierarchical pairs (e.g., ``blue'' and ``blue jacket'') exhibit different embedding magnitudes.
Magnitude differences between $e_i^{+}$ and $e_{i+1}^{+}$ distort the exterior angle $\Xi$ and produce inconsistent hierarchical estimation. 
We normalize the embeddings to remove magnitude variation and preserve the relative geometry of hierarchical representations. Angular-only supervision induces directional bias and degrades representational fidelity; therefore, embedding-space regularization preserves the pretrained knowledge.\\
\textbf{Local-reference entailment contrast (Fig.~\ref{fig:loss_problem} Right)}. Existing contrastive learning is formulated using exterior angles, where sentence pairs are trained with respect to a fixed reference-conditioned point. 
A key consideration in sentence-level contrastive learning with a text encoder is that a negative $e_{t}^{-}$ is also positive with respect to the previous tier $e_{t-1}^{+}$.
This creates an inconsistent hierarchy in which the same sample can be treated as negative at the current tier but positive at the previous tier.
To address this hierarchical inconsistency, we use both a global reference (Global ref.) and a dynamic local reference (Local ref.), positioning samples in opposite directions with respect to the local reference in $r$.

\noindent\textbf{Hierarchy entailment objective.}
Let $l$ denote the tier of HiVG in the \textnormal{object-attribute-relation} hierarchy.
Let $e_t$ be an embedding at tier $t$ of HiVG, $e_{t}^{+}$ its positive counterpart at tier $l$, and $e_{t}^{-}$ a negative sample at the same tier.
We use two complementary losses—(i) alignment across tiers and (ii) within-tier discrimination—defined compactly as
\begin{equation}
\resizebox{0.88\columnwidth}{!}{$
\begin{array}{c}
\mathcal{L}_{\text{H}^{+}}
= \sum_{t=1}^{l} \Xi \langle e_t^{+}, e_{t+1}^{+} \rangle
+ \Xi \langle e_t^{+}, e_{t+1}^{-} \rangle, \\[3pt]

\mathcal{L}_{\text{H}^{-}}
= \sum_{t=1}^{l} \Xi \langle e_t^{+}, e_t^{-} \rangle, \\[3pt]

b' = s(b-r)-a', \qquad
s = 1\ (H^{+}),\ -1\ (H^{-}),
\end{array}$}
\label{eq:hier_objectives}
\end{equation}
with normalization $e = \frac{e - r}{\| e - r \| + \epsilon}, \quad e \in \{e^{+}, e^{-}\}$ for geometry-preserving entailment.
$r$ is adapted to expose directional differences not captured by global alignment and to account for locally meaningful variation.
The proposed formulation minimizes directional deviation from the reference point, which helps preserve the intrinsic structure of the embedding space while enforcing separation between negative pairs.

\subsubsection{Training objectives}
For learning contextualized features, the objective function of hierarchical and disentangled representation learning, $\mathcal{L}_{\text{TaSe}}$, is defined as:
\begin{equation}
    \mathcal{L}_{\text{TaSe}} = \mathcal{L}_{\text{\disenloss}}+ \mathcal{L}_{\text{H}^{+}} + \mathcal{L}_{\text{H}^{-}}.
    \label{eq:re}
\end{equation}
The loss function updates the LoRA parameters of the text encoder and the TriDe module as follows:
\begin{equation}
    \mathcal{L} = \mathcal{L}_{class} + \mathcal{L}_{bbox} +  \mathcal{L}_{giou} + \mathcal{L}_{\text{TaSe}},
\end{equation}
where $\mathcal{L}_{class}$ represents Focal loss \citep{lin2017focal}, $\mathcal{L}_{bbox}$ represents L1 loss, and $\mathcal{L}_{giou}$ represents generalized intersection over union (GIoU) loss \citep{rezatofighi2019generalized}.

\section{Experiments}
\label{sub:evaluation}
This section compares our method with baselines.
The following sections provide the implementation details (Sec. \ref{subsec:setup_main}), the main results for performance comparison (Sec. \ref{subsec:main_results}), and ablation studies conducted to analyze the results in three benchmark datasets (Sec. \ref{subsec:ablation}).

\subsection{Experimental settings}
\label{subsec:setup_main}
\textbf{Implementation details.}
We build our method based on GLEE \citep{glee}, a pre-trained foundation model composed of MaskDINO \citep{li2023mask} and CLIP \citep{clip} text-image encoders.
GLEE was selected as a baseline because it is a powerful vision–language foundation model that performs well on many benchmarks (e.g., RefCOCO \citep{yu2016modeling}), yet still faces challenges in contextualizing text embeddings.
We provide more details of the experimental settings in Appendix~\ref{sec:experimental_setting}.

\noindent\textbf{Benchmarks and evaluation metrics.}
We evaluate language-based object detection on four benchmarks in a \textbf{phrase-level zero-shot setting}.
1) D$^{3}$ \citep{d3} dataset is a widely used benchmark for visual grounding tasks.
The dataset includes negative instances, multi-target scenarios, and long sentences.
2) OmniLabel \citep{omnilabel} dataset is an open-vocabulary detection dataset including both positive and negative descriptions of spatial relations, actions, and numeracy. 
Additionally, we evaluate on gRefCOCO~\cite{he2023grec} and ODinW~\cite{li2022elevater} to assess performance on referring expression tasks.
We use mean average precision (mAP) to validate the language-based object detection task.

\subsection{Main Results}
\label{subsec:main_results}
We investigate the impact of object detection on disentanglement and hierarchical representation learning through a set of analyses.
\begin{tcolorbox}[colback=gray!8,colframe=gray!40,boxrule=0.3pt,arc=2pt,
left=4pt,right=4pt,top=2pt,bottom=2pt]
\footnotesize
\textbf{Key Findings.}
Hierarchical text representation learning mitigates sentence-level ambiguity and improves linguistic compositionality for vision-language alignment.
\textbf{(1)} Hierarchical supervision improves compositionality by structuring sentence embeddings with entailment relations (Tab.~\ref{tab:hierarchy}); and
\textbf{(2)} disentangling shared context and token-specific semantics supports hierarchical learning (Tab.~\ref{tab:disentangle}).
\end{tcolorbox}

\setlength{\tabcolsep}{4pt}
\begin{table}[t]
\centering
\resizebox{\linewidth}{!}{
\begin{tabular}{l *{6}{c}}
\toprule
\multirow{2}{*}{Model} 
& \multicolumn{3}{c}{D$^3$ (default)} 
& \multicolumn{3}{c}{OmniLabel (default)} \\
\cmidrule(r){2-4} \cmidrule(l){5-7}
 & FULL & PRES & ABS
 & AP & AP$_c$ & AP$_d$ \\
\midrule
OFA-L~\citep{wang2022ofa} & 4.2 & 4.1 & 4.6 & 2.7 & 2.7 & 2.6 \\
MDETR~\citep{mdetr} & - & - & - & - & 4.7 & 9.1 \\
OWL~\citep{owl} & 9.6 & 10.7 & 6.4 & 8.0 & 15.6 & 5.4 \\

UNINEXT~\citep{uninext} & 21.6 & 23.7 & 15.4 & 22.2 & 27.2 & 18.8 \\
G-DINO~\citep{gdino} & 20.7 & 20.1 & 22.5 & 19.3 & 23.6 & 16.4 \\
GEN~\citep{gen} & 21.4 & 20.6 & 23.7 & 22.2 & 27.2 & 18.8 \\

GLIP~\citep{glip} & 19.1 & 18.3 & 21.5 & 19.3 & 23.6 & 16.4 \\
\rowcolor{gray!15} GLEE-Lite$^{*}$~\citep{glee} & 27.6 & 26.8 & 30.1 & 21.7 & 36.6 & 15.4 \\
GLIP + \desco~\citep{desco} & 24.2 & 22.9 & 27.8 & 23.8 & 27.4 & 21.0 \\
GLEE-Lite + \desco & 28.3 & 27.6 & 30.3 & 24.6 & \textbf{37.3} & 18.3 \\
\rowcolor{yellow!20}
GLEE-Lite + TaSe & \textbf{30.7} & \textbf{29.9} & \textbf{33.2} & \textbf{26.9} & 36.8 & \textbf{21.2} \\
\bottomrule
\end{tabular}}
\vspace{-2mm}
\caption{Evaluation on D$^3$ and OmniLabel.
Performance comparisons with MLLM-based baselines are reported in Tab.~\ref{tab:results_details}.}
\vspace{-2mm}
\label{tab:results}
\end{table}
\begin{table}[t]
\renewcommand\arraystretch{1.05}
\small
\centering
\setlength{\tabcolsep}{3mm}
\resizebox{\linewidth}{!}{
\begin{tabular}{lcccc}
\toprule
Methods & val & testA & testB & ODinW \\
\midrule

MCN~\cite{luo2020multi} & 28.0 & 32.3 & 26.8 & - \\
VLT~\cite{ding2021vision} & 36.6 & 40.2 & 30.2 & - \\
MDETR & 42.7 & 50.0 & 36.5 & 25.1 \\
UNINEXT$^\dagger$~\cite{uninext} & 58.2 & 46.4 & 42.9 & - \\

\rowcolor{gray!15} GLEE-Lite \citep{glee} & 66.7 & 73.0 & 61.8 & 44.4 \\

\rowcolor{yellow!20}
GLEE-Lite + TaSe & \textbf{69.6} & \textbf{76.6} & \textbf{67.7} & \textbf{46.9} \\

\bottomrule
\end{tabular}}

\vspace{-2mm}
\caption{Zero-shot evaluation results on gRefCOCO (val, testA, testB) and the average performance across 13 datasets in ODinW (see Sec. \ref{appendix:detailed_qualitative_results} for detailed results).}
\vspace{-2mm}
\label{tab:grefcoco_odinw}
\end{table}
\begin{table}[t]
\centering

\resizebox{\linewidth}{!}{
\begin{tabular}{@{} lcc}
\toprule
& D$^3$ & OmniLabel\\ 
\midrule

\rowcolor{gray!10}\multicolumn{3}{c}{Where-to-apply disentanglement}\\
\addlinespace[0.2ex]
\cdashline{1-3}
\addlinespace[0.2ex]

w/o disentangling & 27.8\ \textcolor{red}{(+1.0)} & 26.2\ \textcolor{red}{(+4.5)}  \\

\rowcolor{yellow!20}
Token-level disentangling & \textbf{30.7\ \textcolor{red}{(+3.1)}} & \textbf{26.9\ \textcolor{red}{(+5.2)}}  \\

$\llcorner$ Identity initialization & 30.7\ \textcolor{red}{(+3.1)} & 26.9\ \textcolor{red}{(+5.2)} \\

$\llcorner$ Uniform initialization & 28.8\ \textcolor{red}{(+1.2)} & 26.4\ \textcolor{red}{(+4.4)}\\

After pooling & 28.6\ \textcolor{red}{(+1.4)} & 22.6\ \textcolor{red}{(+0.9)}  \\

\midrule

\rowcolor{gray!10}\multicolumn{3}{c}{How-to-apply disentanglement}\\
\addlinespace[0.2ex]
\cdashline{1-3}
\addlinespace[0.2ex]

Self-attention & 29.0\ \textcolor{red}{(+1.4)} & 25.5\ \textcolor{red}{(+3.8)}\\

Learnable query & 29.6\ \textcolor{red}{(+2.0)} & 24.7\ \textcolor{red}{(+3.0)} \\

\rowcolor{yellow!20}
Learnable key \& value & \textbf{30.7\ \textcolor{red}{(+3.1)}} & \textbf{26.9\ \textcolor{red}{(+5.2)}} \\

\midrule

\rowcolor{gray!10}\multicolumn{3}{c}{Effectiveness of disentangling components (\# of learnable vector)}\\
\addlinespace[0.2ex]
\cdashline{1-3}
\addlinespace[0.2ex]

1 (w/o disentanglement) & 29.4\ \textcolor{red}{(+1.8)} & 25.4\ \textcolor{red}{(+3.7)} \\

2 (Object + Attribute) & 29.5\ \textcolor{red}{(+1.9)} & 26.5\ \textcolor{red}{(+4.8)} \\

\rowcolor{yellow!20}
3 (Object + Attribute + Relation) & \textbf{30.7\ \textcolor{red}{(+3.1)}} & \textbf{26.9\ \textcolor{red}{(+5.2)}} \\

\bottomrule
\end{tabular}}
\vspace{-2mm}
\caption{Ablation studies on disentangled representations (Sec.~\ref{subsec:drl}) with hierarchy entailment.}
\label{tab:disentangle}
\vspace{3mm}

\resizebox{0.95\linewidth}{!}{
\begin{tabular}{@{} lcc}
    \toprule 
    & D$^3$ & OmniLabel\\ 
    \midrule

    \rowcolor{gray!15} 
    GLEE-Lite \cite{glee} 
    & 27.6 
    & 21.7 \\

    + LoRA with HiVG data
    & 27.5\ \textcolor{red}{(-0.1)} 
    & 21.7\ \textcolor{red}{(+0.0)} \\

    \specialrule{0.08em}{0.4ex}{0.4ex}

    \rowcolor{gray!10}
    \multicolumn{3}{c}{Comparison of contrastive learning objectives with LoRA}\\
    \cdashline{1-3}
    \addlinespace[0.6ex]

    $\;$ + $\mathcal{L}_{\text{CL}}$ 
    & 26.9\ \textcolor{red}{(-0.7)} 
    & 23.9\ \textcolor{red}{(+2.2)} \\

    $\;$ + $\mathcal{L}_{\text{RE}}$
    & 27.5\ \textcolor{red}{(-0.1)} 
    & 25.7\ \textcolor{red}{(+4.0)} \\

    \rowcolor{yellow!20} 
    $\;$ + $\mathcal{L}_{\text{H}^{\pm}}$ in Eq.~\ref{eq:hier_objectives} 
    & 28.6\ \textcolor{red}{(+1.0)} 
    & \textbf{26.2\ \textcolor{red}{(+4.5)}} \\

    $\quad \;\;\llcorner$ w/o $\mathcal{L}_{\text{H}^{-}}$ 
    & \textbf{28.8\ \textcolor{red}{(+1.2)}} 
    & 24.8\ \textcolor{red}{(+3.1)} \\

    $\quad \;\;\llcorner$ w/o $\mathcal{L}_{\text{H}^{+}}$ 
    & 27.7\ \textcolor{red}{(+0.1)} 
    & 23.6\ \textcolor{red}{(+1.9)} \\

    $\;$ + Reverse objective: -$\mathcal{L}_{\text{H}^{\pm}}$ 
    & 26.7\ \textcolor{red}{(-0.9)} 
    & 22.1\ \textcolor{red}{(+0.4)} \\

    \specialrule{0.08em}{0.4ex}{0.4ex}

    \rowcolor{gray!10}
    \multicolumn{3}{c}{TriDe module ablation with LoRA on HiVG data}\\
    \cdashline{1-3}
    \addlinespace[0.6ex]

    \rowcolor{yellow!20} 
    $\;$ + TriDe module with $\mathcal{L}_{\text{TriDe}}$
    & \textbf{28.8\ \textcolor{red}{(+1.2)}} 
    & 25.8\ \textcolor{red}{(+4.1)} \\

    \bottomrule
\end{tabular}
}

\vspace{-2mm}
\caption{Ablation studies on hierarchy objective and the TriDe module. $\mathcal{L}_{\text{CL}}$ and $\mathcal{L}_{\text{RE}}$ represent contrastive loss \citep{oord2018representation} and the RE objective, respectively.}
\vspace{-4mm}
\label{tab:hierarchy}
\end{table}
\begin{table}[t]
\centering
\resizebox{\linewidth}{!}{
\begin{tabular}{lccccc}
\toprule
\textbf{Model} & \textbf{Params} & \textbf{FULL}$\uparrow$ & \textbf{PRES}$\uparrow$ & \textbf{ABS}$\uparrow$ & \textbf{Latency}$\downarrow$ \\
\midrule
Qwen3-VL-2B-Instruct & 2B & 16.0 & 17.0 & 11.2 & 582 \\
Qwen3-VL-4B-Instruct & 4B & 24.3 & 25.8 & 19.7 & 1,791 \\
InternVL2-8B-Instruct & 8B & 9.8 & 11.0 & 6.2 & 932 \\
\rowcolor{gray!10}
\textbf{GLEE + TaSe (Ours)} & \textbf{0.18B} & \textbf{30.7} & \textbf{29.9} & \textbf{33.2} & \textbf{131} \\
\bottomrule
\end{tabular}
}
\vspace{-2mm}
\caption{Comparison with MLLM-based baselines on D$^3$. Latency is measured in milliseconds.}
\vspace{-6mm}
\label{tab:mllm_latency}
\end{table}

\noindent\textbf{Zero-shot generalization performance.}
The proposed model improves upon the baseline while adapting only the text embeddings (Tab.~\ref{tab:results}).
Compared to GLEE, which served as the vision foundation model, we observe improvements of $+3.1$ in D$^3$ and $+5.2$ in OmniLabel AP scores.
The OmniLabel results show that hierarchical learning improves performance in zero-shot evaluation, and the gains observed on open-vocabulary benchmarks further demonstrate its effectiveness.
We also compare TaSe with baseline models on the gRefCOCO and ODinW in Tab.~\ref{tab:grefcoco} and \ref{tab:odinw}.
TaSe achieves an average improvement of $+4.9$ AP on ODinW and improves object detection performance with textual queries on gRefCOCO.

\noindent\textbf{Does hierarchical learning truly improve linguistic reasoning?}
We examine whether hierarchical learning of sentence embeddings improves linguistic compositionality.
To evaluate the methodology, we compare our method with DesCo~\cite{desco}, a caption augmentation method.
We randomly sample sentences from HiVG for the DesCo configuration. The selected sentence is concatenated with the original caption in either order 
($c~\texttt{[SEP]}~c^{-}$ or $c^{-}~\texttt{[SEP]}~c$),
where $c$ and $c^{-}$ denote the original and negative captions, respectively. The augmented components are pooled separately and then averaged.
The \desco improves the GLEE model by $+2.4$ AP on D$^3$ and $+2.3$ AP on OmniLabel.

\begin{figure*}
    \centering
    \begin{minipage}{0.68\textwidth}
        \centering
        \includegraphics[width=\linewidth]{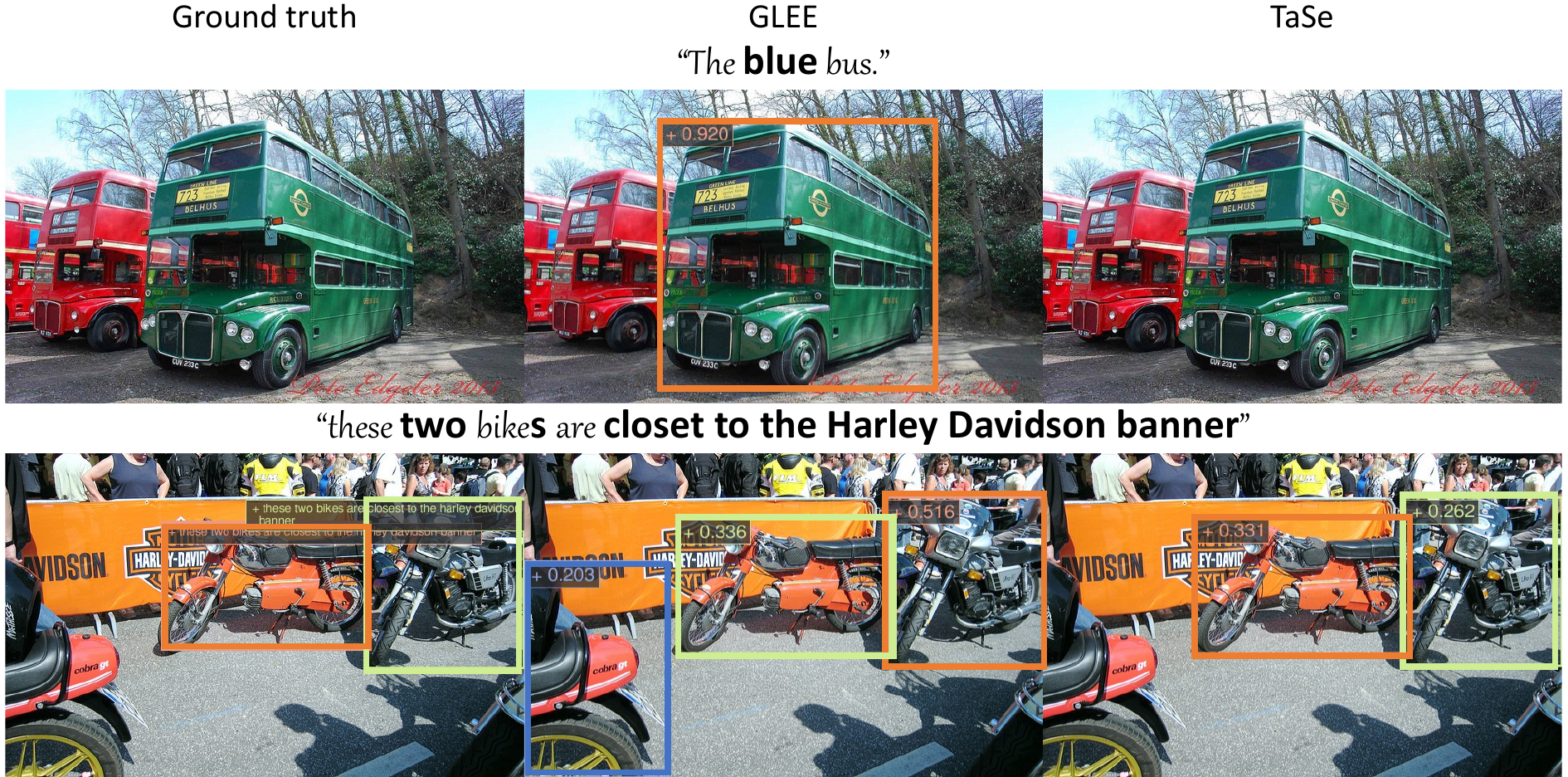}
        \vspace{-2em}
        \caption{Qualitative analysis on OmniLabel. We compare the baseline and \model on language descriptions with attributes and relations.
        }
        \label{fig:main_quality}
        \end{minipage}%
        \hfill
    \begin{minipage}{0.31\textwidth}
    \centering
    \footnotesize
    \setlength{\tabcolsep}{1.8pt}
    \renewcommand{\arraystretch}{1.13}

    \begin{tabularx}{\textwidth}{
        >{\centering\arraybackslash}p{1.5cm}|
        >{\centering\arraybackslash}p{1.0cm}|
        >{\centering\arraybackslash}X}
    \hline
    Measure & Base & +TaSe \\
    \hline

    \multicolumn{3}{c}{\textbf{Grounding DINO-B~\scriptsize{\cite{gdino}}}} \\
    D$^3$
        & 20.7
        & \mbox{\textbf{24.3} {\color{red}(+3.6)}} \\
    OmniLabel
        & 19.3
        & \mbox{\textbf{28.6} {\color{red}(+7.3)}} \\
    Latency
        & 114
        & \mbox{123 {\color{blue}(+9)}} \\
    \hline

    \multicolumn{3}{c}{\textbf{GLIP-T~\scriptsize{\cite{glip}}}} \\
    D$^3$
        & 19.1
        & \mbox{\textbf{21.5} {\color{red}(+2.4)}} \\
    OmniLabel
        & 19.3
        & \mbox{\textbf{19.9} {\color{red}(+0.6)}} \\
    Latency
        & 122
        & \mbox{138 {\color{blue}(+16)}} \\
    \hline

    \multicolumn{3}{c}{\textbf{Qwen3-VL-2B~\scriptsize{\cite{yang2025qwen3}}}} \\
    D$^3$
        & 16.0
        & \mbox{\textbf{41.6} {\color{red}(+25.6)}} \\
    OmniLabel
        & 3.8
        & \mbox{\textbf{30.4} {\color{red}(+26.6)}} \\
    Latency
        & 582
        & \mbox{583 {\color{blue}(+1)}} \\
    \hline
    \end{tabularx}

    \vspace{-1em}
    \captionof{table}{Generalization and latency of TaSe across detectors}
    \label{tab:generalization}
\end{minipage}
\end{figure*}

\noindent\textbf{Qualitative results.}
We present two qualitative examples in Fig.~\ref{fig:main_quality} to illustrate the effectiveness of our hierarchical entailment learning.
The first case shows negative cases containing an attribute (i.e.,  blue), and the second case presents a positive case with attributes and descriptive relations (e.g., numeracy and text in the image).
In the first case, GLEE incorrectly assigns a confident score of $0.92$ to the language query.
In the second case, GLEE predicts all bikes as positives, including those that do not correspond to the queried bike.
In contrast, \model captures contextual information related to category names, and hierarchical entailment helps reduce false positives.

\subsection{Ablation Studies}
\label{subsec:ablation}

\textbf{Is negative caption augmentation alone sufficient?}
We ablate the hierarchical entailment loss to compare learning with positive and negative pairs. Fine-tuning with positives improves OmniLabel by $+3.1$ AP, while negatives yield $+1.9$ AP. 
We further conduct the objective and observe performance degradation when positives and negatives are aligned in reversed directions (Tab.~\ref{tab:hierarchy}). 
Hierarchical entailment improves sentence-level embeddings and strengthens semantic alignment with visual representations and performance.
While prior captioning methods \cite{desco, doveh2023dense} focus on negatives, our approach achieves further performance gains via positive hierarchy.

\noindent\textbf{What advantages does our hierarchical loss offer over traditional contrastive loss?}
In Tab.~\ref{tab:hierarchy}, we conduct an ablation study to validate the effectiveness of our hierarchical loss. 
Within the base setting (GLEE with LoRA), we evaluate three configurations: (1) conventional contrastive loss ($\mathcal{L}_{\text{CL}}$); (2) the RE objective ($\mathcal{L}_{\text{RE}}$); and (3) the proposed method. 
For sentence-level hierarchical aggregation, our loss $\mathcal{L}_{\text{H}}$ outperforms contrastive baselines. 
Our reference-based hierarchical induction aligns them hierarchically and improves sentence-level meaning and performance.

\noindent\textbf{Ablation on where and how to disentangle in text representation.}
Tab.~\ref{tab:disentangle} reports ablation studies analyzing the design choices of the \disenloss module.
The results show that the disentanglement location plays a key role in controlling the granularity of the representation.
We compare three modes for constructing compositional text embeddings: (1) no disentanglement, (2) token-level disentanglement (before pooling), and (3) disentangled embeddings after pooling. Preserving text positional bias during token-level disentanglement improves performance and generalization. For the design of the \disenloss module, we further examine three attention variants: (1) standard self-attention, (2) learnable queries, and (3) key–value configurations. The key-value setting improves performance by using keys to identify component types and values to encode the corresponding semantic contents.

\noindent\textbf{Is it beneficial to disentangle the representation into three components?}
Conventional detectors \citep{desco,yuksekgonul2023when} disentangle objects and attributes, whereas we separate representations into three components—object, attribute, and relation—and evaluate their effectiveness. 
As shown at the bottom of Tab.~\ref{tab:disentangle}, overall, the performance of three-component disentanglement is higher than two-component (i.e., object and attribute) disentanglement.
Three-component disentanglement introduces an inductive bias for complex linguistic structures, making longer sentences more robust to negative samples.
Finer-grained text embeddings better expose structural information in representations, motivating dataset composition criteria based on objects, attributes, and relations. We provide the disentangled embedding results in Appendix~\ref{sec:exterior_angle_analysis}.

\noindent\textbf{Generalization performance across diverse detector architectures.}
Experiments with GroundingDINO \cite{gdino}, GLIP \cite{glip}, and Qwen3-VL-2B \cite{yang2025qwen3} are conducted to validate the generalizability of TaSe. 
The results show consistent performance improvements (Tab.~\ref{tab:generalization}).
The proposed approach improves performance on D$^3$ (FULL AP) and OmniLabel (AP$_d$) with only marginal latency overhead. Details of the learning setup are provided in Appendix Sec.~\ref{sec:detectors_setup}.
\model improves compositional reasoning and robust VL alignment in the shared embedding space and generalizes across model architectures.

\noindent\textbf{Efficiency Comparison with MLLM Baselines.}
Recent MLLMs have shown strong zero-shot generalization ability \cite{yang2025qwen3, internvl}.
We compare inference latency with Qwen3-VL \cite{yang2025qwen3} and InternVL2 \cite{internvl} on 100 samples to assess the efficiency of our model against MLLM baselines (see Tab.~\ref{tab:mllm_latency}).
Compared with these MLLM baselines, our model requires only 0.18B parameters and achieves substantially lower latency, demonstrating higher efficiency for detection-oriented inference. 
We emphasize the need to develop efficient detectors for real-time detection tasks.

\noindent\textbf{Additional analyses on HiVG.} We report the performance of TaSe on HiVG datasets generated using different LLMs in Appendix~\ref{appendix:caption_llms}, and provide a human evaluation of HiVG in Appendix~\ref{appendix:caption_eval}.

\section{Conclusion}
\label{sub:conclusion}
This study proposed a disentanglement and hierarchical aggregation framework for constructing contextualized sentence representations within language-based object detection.
Additionally, we generated re-captioned data for object detection using hierarchy concepts.
\model improved linguistic compositionality, which serves as a key learning factor and leads to competitive results.
The results indicated that hierarchical entailment allows learning the granularity of text embedding to distinguish descriptive sentences.
This study highlights the need for further exploration of the underlying linguistic compositionality in future studies for downstream vision tasks.

%-------------------------------------------------------------------------

\section*{Limitations}
This section outlines the limitations of our proposed method. 
We propose a disentanglement and hierarchical aggregation framework to enhance linguistic compositionality in VLMs. 
The three disentangled components (i.e., object–attribute–relation) may not represent the optimal separation of language semantics. Despite this limitation, we adopt this decomposition to analyze the behavior of the embedding space. Our analysis shows that the proposed objective induces compositional structure and semantic disentanglement in the learned representations.  In the detection task we consider, disentangling objects, attributes, and relations in the embedding space reduces false positives and improves performance.

\section*{Ethics Statements}
% code ethics
As for VL-based detection models, we access all open-source LMs via the Hugging Face hub \cite{wolf2020transformers}. The number of parameters is presented in Tab.~\ref{tab:configuration}. All associated licenses permit user access for research purposes, and we have agreed to follow all terms of use.
We validate the effectiveness of our proposed approach using the
D3 \cite{d3}, OmniLabel \cite{omnilabel}, GRefCOCO \cite{he2023grec}, and ODinW \cite{glip} reference datasets. These datasets are sourced from publicly available resources on legitimate websites and do not contain any sensitive data. 

% ours data ethics
We re-captioned our dataset, utilizing the publicly available Llama 3 \cite{dubey2024llama}, GPT-4-nano \cite{achiam2023gpt4}, and Qwen3 \cite{yang2025qwen3} models released on the Hugging Face hub \cite{wolf2020transformers}.
Additional statistics and details of the dataset are presented in Appendix~\ref {sec:data_details}.

During the preparation of this work, the authors used ChatGPT \citep{gpt5} in order to improve readability. After using this tool/service, the author reviewed and edited the content as needed and takes full responsibility for the content of the publication.

\section*{Acknowledgements}
This research was supported by the Institute of Information \& communications Technology Planning \& Evaluation (IITP) grant funded by the Korea government(MSIT) (No. RS-2019-II190079, Artificial Intelligence Graduate School Program(Korea University), 1\%; No. RS-2025-25439490, 29\%), the KOCCA grant (RS-2024-00345025, 10\%), the National Research Foundation of Korea(NRF) grant funded by the Korea government(MSIT)(No. RS-2024-00341514, 20\%; No. RS-2025-02263628, 20\%; No. RS-2026-25499205, 20\%)
%-------------------------------------------------------------------------

\bibliography{manuscript}

\clearpage

\appendix

\noindent
For a comprehensive understanding of the TaSe and HiVG dataset, we provide this Appendix. The following table of contents gives 
a concise overview and directs readers to specific sections of interest.

\vspace{1em}

{\Large \textbf{Contents}}

\vspace{0.5em}

\begin{tabular}{p{0.89\linewidth}r}

\textbf{A. HiVG Dataset Construction} \dotfill 13 \\
\hspace{0.5em} A.1 Data Generation Pipeline \dotfill 13 \\
\hspace{0.5em} A.2 Dataset Statistics \dotfill 14 \\
\hspace{0.5em} A.3 Performance Comparison based on Captions Generated by Different LLMs \dotfill 14 \\
\hspace{0.5em} A.4 Human Evaluation for Generated Captions \dotfill 15 \\
\vspace{0.2em}
\textbf{B. Experimental Setup} \dotfill 15 \\
\hspace{0.5em} B.1 Implementation Details \dotfill 15 \\
\hspace{0.5em} B.2 Model Size and Budget \dotfill 15 \\
\hspace{0.5em} B.3 Experimental Setup for Additional Detectors \dotfill 18 \\
\vspace{0.2em}
\textbf{C. Additional Results} \dotfill 18 \\
\hspace{0.5em} C.1 Detailed Quantitative Results \dotfill 18 \\
\hspace{0.5em} C.2 Additional Qualitative Results \dotfill 19 \\
\hspace{0.5em} C.3 Additional Ablation Studies \dotfill 19 \\
\hspace{0.5em} C.4 Pseudo Code for TaSe \dotfill 19 \\
\vspace{0.2em}
\textbf{D. Discussions} \dotfill 19 \\
\hspace{0.5em} D.1 How are embeddings structured after disentanglement? \dotfill 19 \\
\hspace{0.5em} D.2 Comparative Analysis of Text Embeddings across State-of-the-Art Vision-Language Models \dotfill 19 \\
\hspace{0.5em} D.3 Exterior Angle Analysis \dotfill 23 \\
\hspace{0.5em} D.4  Analysis of Hierarchical Objectives \dotfill 23 \\
\hspace{0.5em} D.5 Failure Cases \dotfill 23\\[0.3em]

\end{tabular}

\section{HiVG Dataset Construction}
\label{sec:data_details}

\subsection{Data Generation Pipeline}
\begin{figure}[t]
    \centering
    \includegraphics[width=\linewidth]{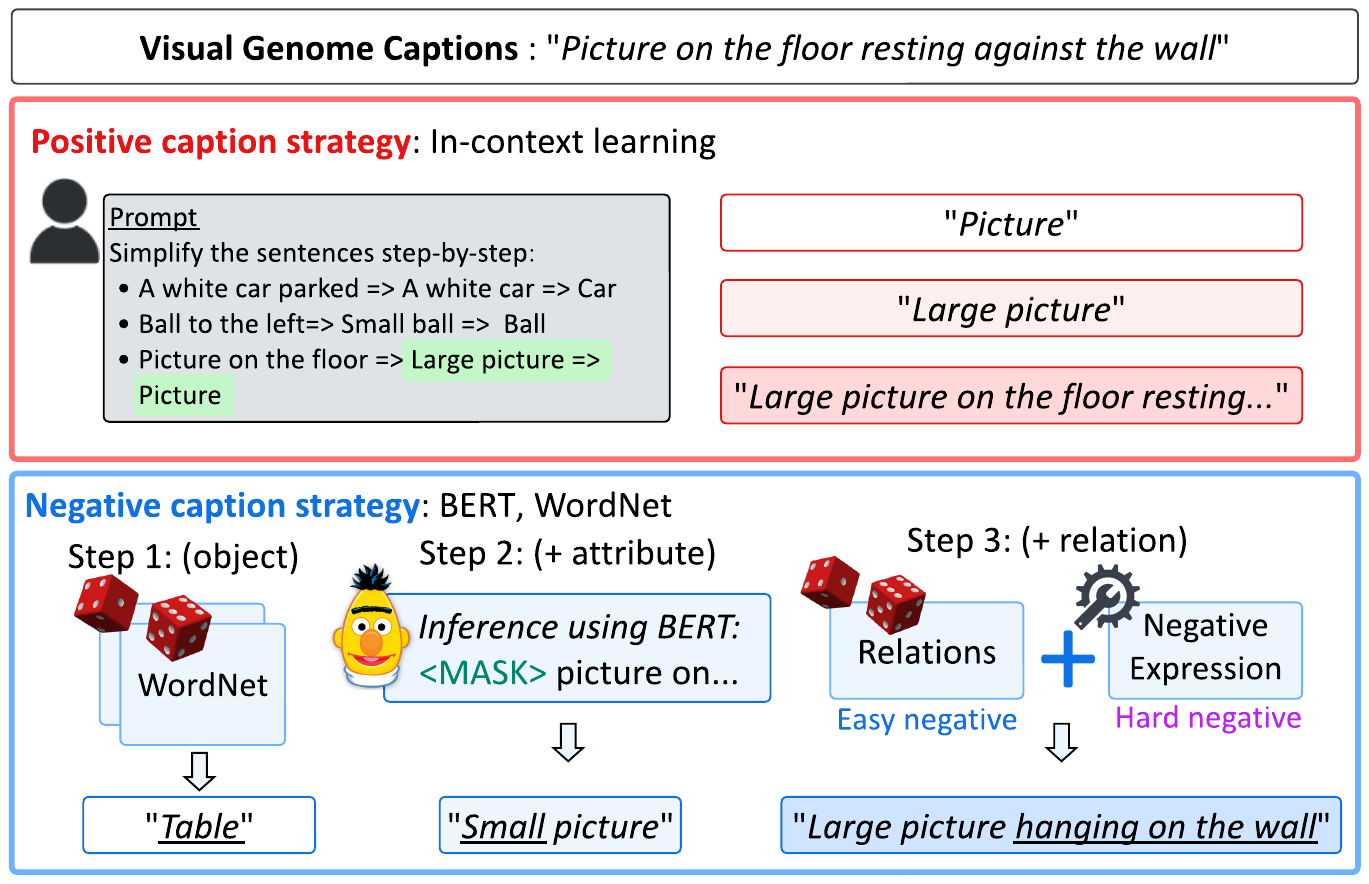}
    \caption{Overview of generating positive and negative captions. Positive captions are derived using in-context learning based on Llama-3 \citep{dubey2024llama}.}
    \label{fig:dataset}
\end{figure}

Our hierarchical captioning pipeline is illustrated in Fig. \ref{fig:dataset}.
%, which highlights key differences from conventional captioning approaches.
We generate a 10 K re-captioned dataset from the Visual Genome dataset.
We first filter out all Visual Genome captions with fewer than six words to ensure sufficient semantic richness. To reduce hallucination, we enforced the hierarchical format ``Tier 3=>Tier 2=>Tier 1'' using the delimiter ``=>''.
Captions that did not conform to this structure are filtered and regenerated.
As shown in Fig.~\ref{fig:llm}, GPT-4 \cite{achiam2023gpt4} data demonstrate the highest performance, with Llama-3~\cite{dubey2024llama} second and Qwen3 \cite{yang2025qwen3}.

\subsubsection{Creating Hierarchical Positive Captions}
\label{sec:pos_captioning}
%\noindent\textbf{Positive tier 1}
For positive captions, we leverage in-context learning using LLaMA \citep{dubey2024llama} to transform the remaining Visual Genome captions into a three-tier hierarchical structure.
To enhance attribute diversity in captions, we draw on common visual concepts \citep{huang2023t2i, lin2024evaluating} to define a set of visual attributes (spatial, color, number, and size) and randomly select one to modify the object's attributes.
Based on our experiments, we set the randomization ratio to 50\%.
To fully align with the in-context learning demonstration format, samples that lack attributes or relations are also incorporated into the learning process.

\subsubsection{Creating Hierarchical Negative Captions}

\noindent\textbf{Negative tier 1.}
One of the challenges in language-based object detection is effectively handling negative samples, which often report higher false negative rates compared to false positives.
To address the issue, we focus on both re-captioning hard and easy negative samples.
For hard negative samples in tier 1, we replace the positive object with an antonym (e.g., \textit{man} is replaced by \textit{woman}) or a random concrete noun. 
Easy negative samples are generated by selecting nouns from ImageNet-1K \citep{deng2009imagenet} classes and lexical databases such as WordNet and ConceptNet.
Additionally, we insert a negative determiner to the object (e.g., \textit{dog} is switched to \textit{no dog}).

\noindent\textbf{Negative tier 2.}
We reuse the same set of visual attributes from the positive captions but replace them with semantically distinct attributes (e.g., \textit{tall building} is replaced by \textit{short building}) to generate hard negative samples. 
We use LLM-based mask-filling \citep{Liu2019RoBERTaAR} to diversify attributes by substituting them with contextually plausible but semantically different terms or by prepending ``not" to create hard negatives (e.g., tall building → not tall building).

\noindent\textbf{Negative tier 3.}
We use a set of common spatial relations (e.g., above and beside) and object-specific relations from the Visual Genome dataset.
These pre-defined object-specific relations ensure that the relation is contextually relevant to the object in question. 
To introduce hard negatives, we apply absence-based transformations by replacing affirmative relations with their negative counterparts (e.g., \textit{with} is replaced by \textit{without}). 
We also leverage LLaMA's \citep{dubey2024llama} sentence completion capability to generate further relation diversity.

Captions that do not adhere to the hierarchical structure (the arrow symbol ``=>'' is missing, or the second tier does not contain the previous tier) are filtered out and regenerated.
By re-captioning using a multi-tiered set of positive and negative captions, our approach is intended to facilitate the learning of hierarchical representations, thereby improving linguistic compositionality.

\begin{figure}
    \centering
    \includegraphics[width=5cm]{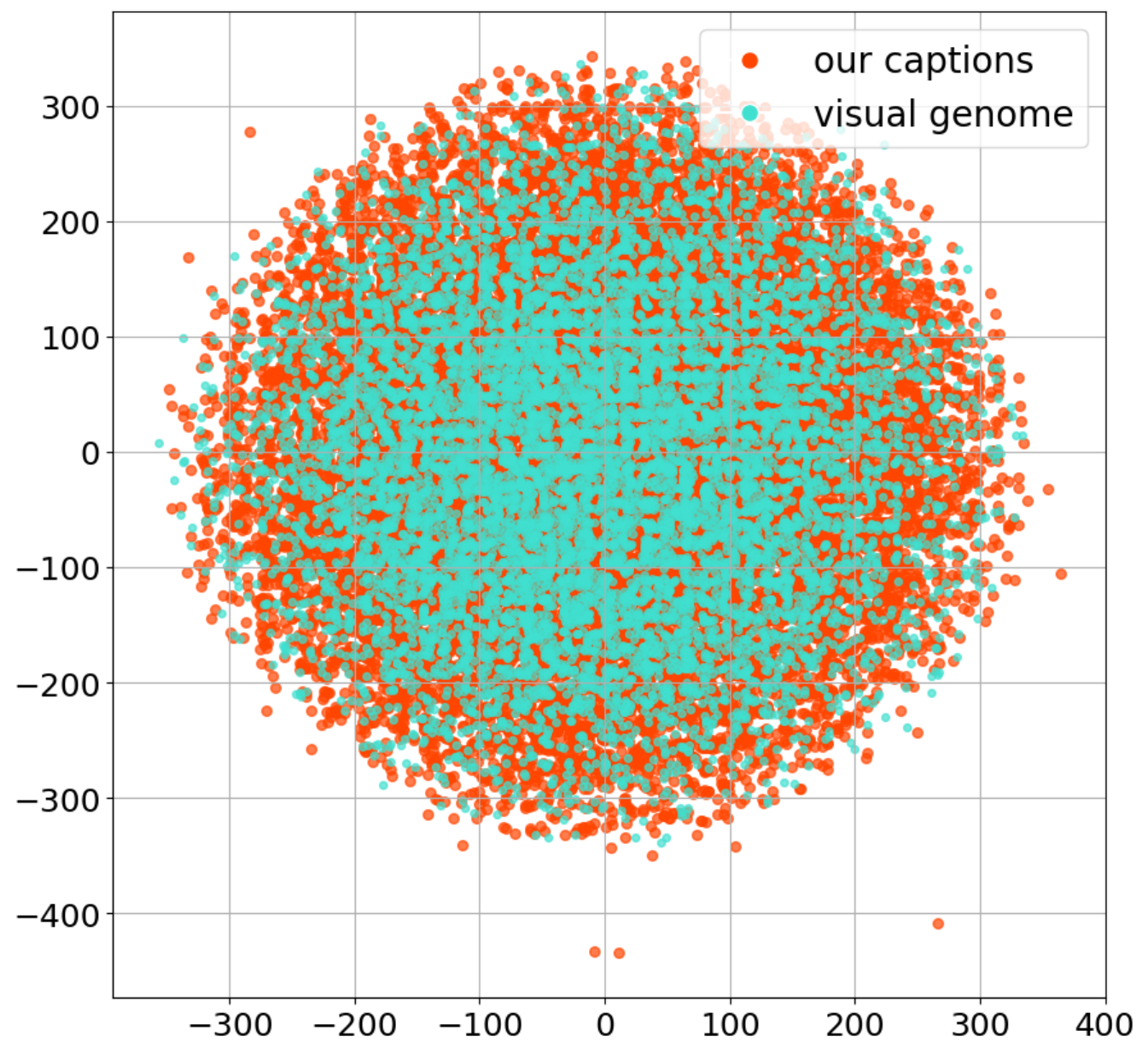}
    \caption{t-SNE plot comparing the clustering of our captions with Visual Genome captions. {\setlength{\fboxsep}{1pt}\colorbox{green!12}{Visual Genome dataset}} and {\setlength{\fboxsep}{1pt}\colorbox{orange!12}{HiVG dataset}} are shown in the plot. Our captions exhibit a broader spread in the 2D space, indicating greater semantic diversity compared to the original Visual Genome captions.}
    \label{fig:data_distribution}
\end{figure}

\subsection{Dataset Statistics}
\begin{figure*}[t]
    \centering
    \includegraphics[width=0.95\linewidth]{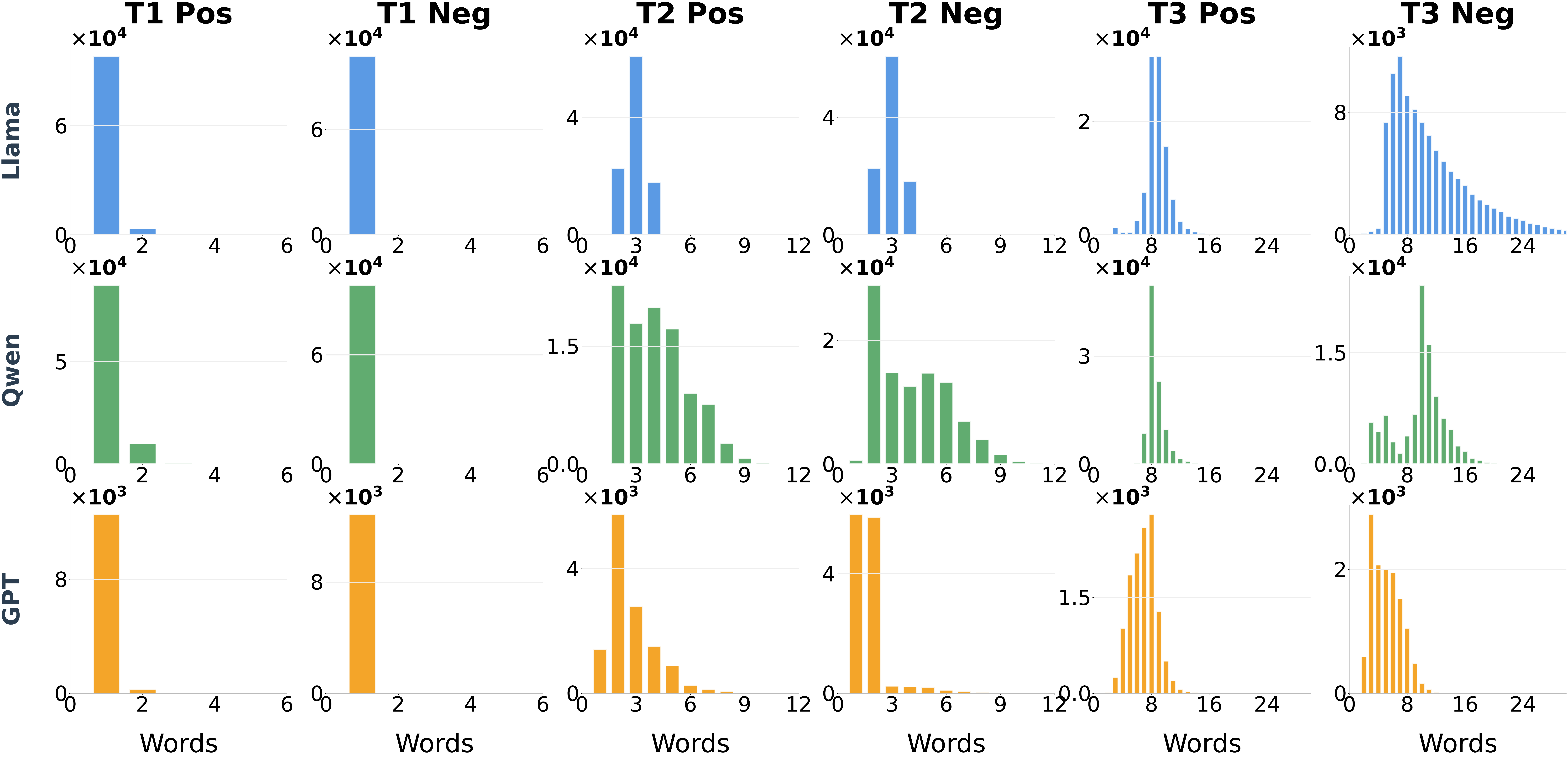}
    \caption{Statistics of the HiVG dataset generated with different LLMs.
    Rows correspond to datasets generated by LLaMA-3-8B, Qwen3-8B, and GPT-4-nano. Columns show the word-count distributions of hierarchical caption tiers, including T1 (category), T2 (attribute), and T3 (relation), each separated into positive (Pos) and negative (Neg) captions.
    The distributions remain consistent across LLMs.}
    \label{fig:vg_statistic}
\end{figure*}

\begin{figure*}
    \centering
    \includegraphics[width=\linewidth]{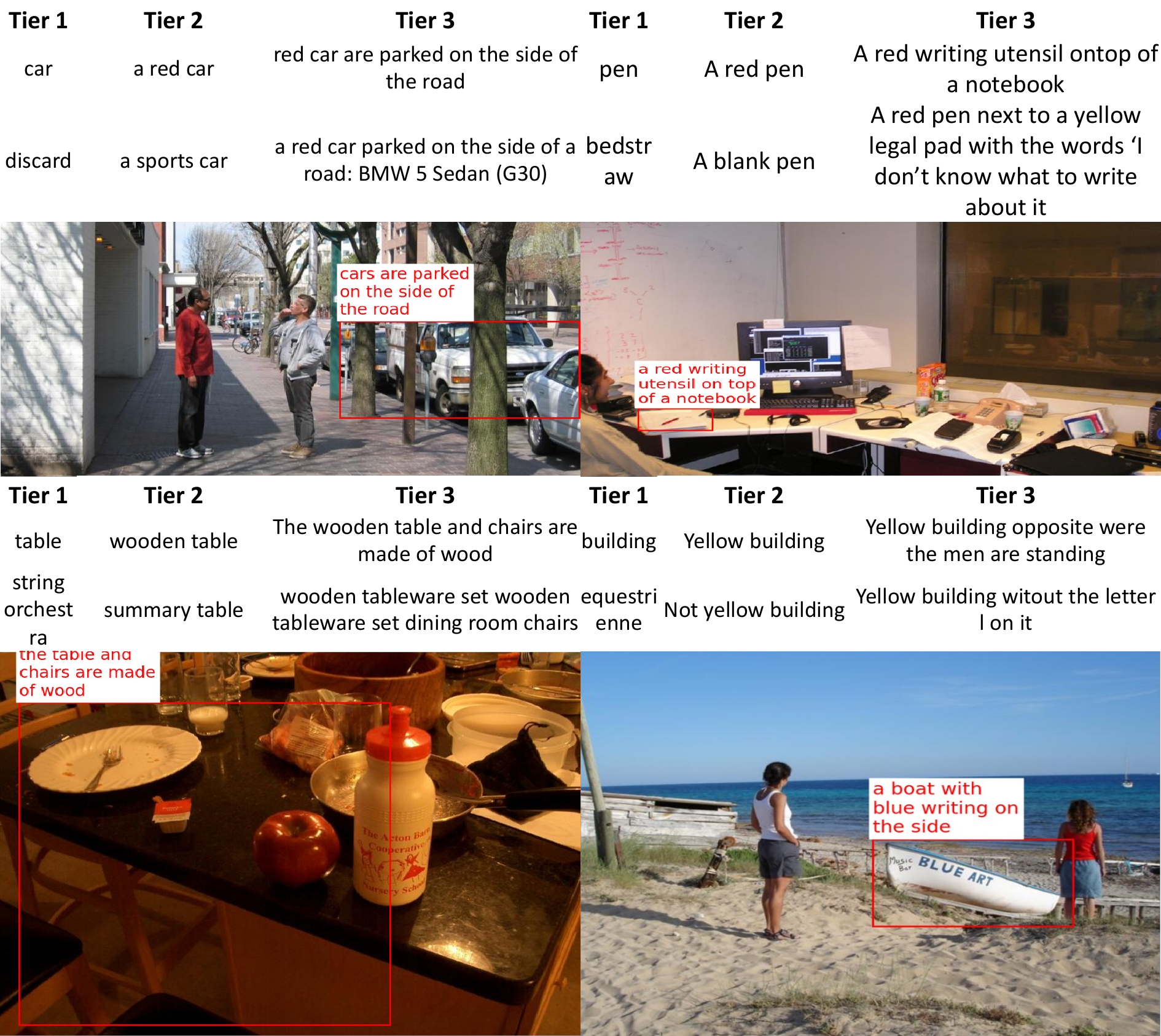}
    \caption{Re-captioning data examples}
    \label{fig:recaptioning_example}
\end{figure*}

\begin{figure}[H]
    \centering
    \includegraphics[width=0.8\linewidth]{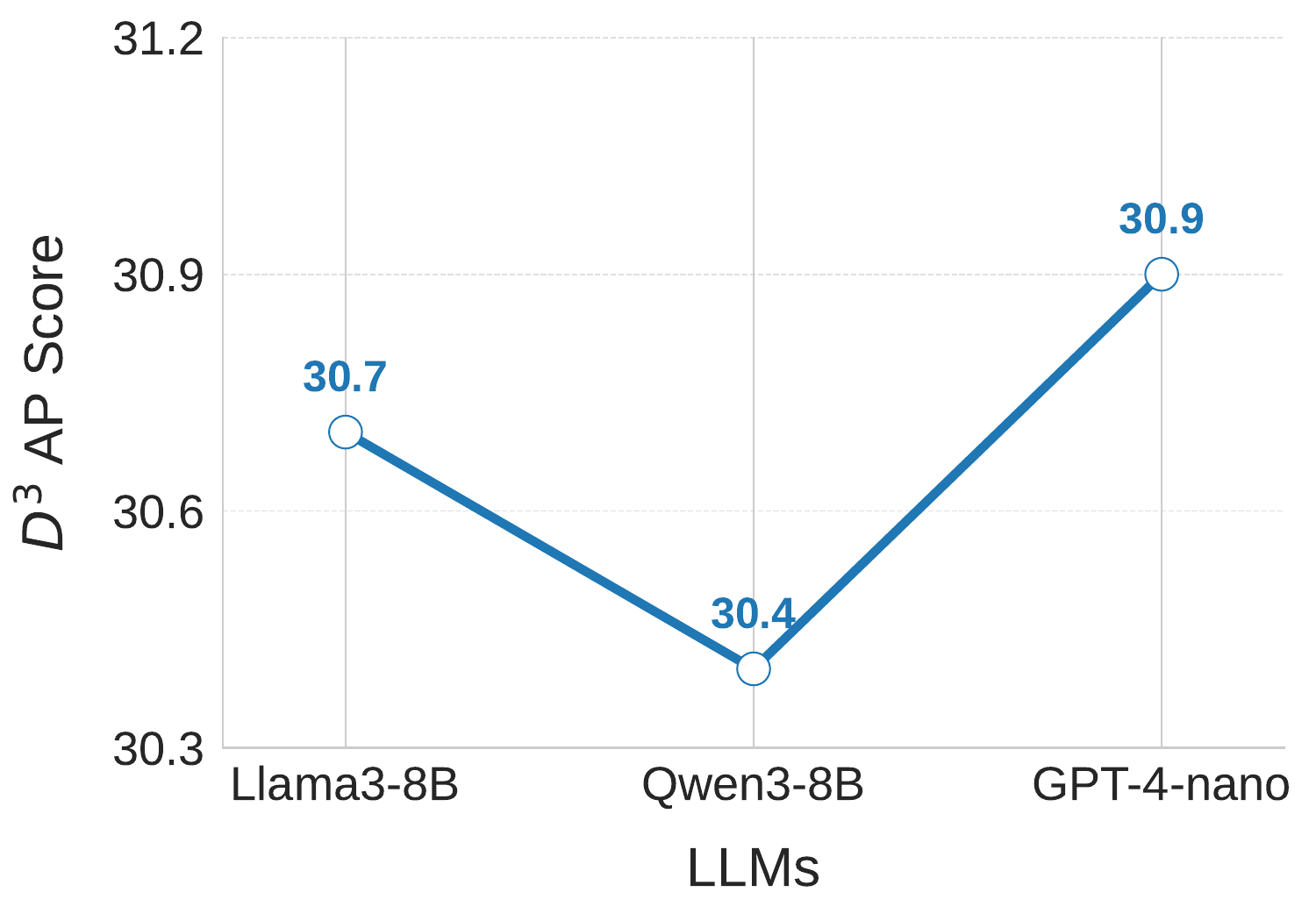}
    \caption{Evaluation of how captions generated by different LLMs affect object detection performance}
    \label{fig:llm}
\end{figure}

The combined dataset consists of 101{,}897 LLaMA-based annotations, with the majority containing $4.8 \pm 5.2$ words, as shown in Fig.~\ref{fig:vg_statistic}.
The Qwen-based dataset contains 98{,}261 annotations with $4.8 \pm 3.9$ words, while the GPT-based dataset includes 101{,}852 annotations with $3.1 \pm 2.6$ words.
GPT generates the shortest captions on average.
Caption length distributions across positive and negative samples remain similar within each tier, with negative captions in the LLaMA and Qwen datasets being slightly longer on average ($\approx$ 1 word).
Tier 3 contains the most linguistically diverse and structurally rich captions, which may benefit semantic reasoning.
We report examples from our hierarchical dataset, HiVG, in Fig. \ref{fig:recaptioning_example}.
Leveraging Visual Genome data for re-captioning, we create a more diverse dataset by incorporating a wider range of classes and the LLM and other datasets.
We also visualize the t-SNE distribution of VisualGenome and HiVG data in Fig.~\ref{fig:data_distribution}. The figure shows that the generated data maintains a distribution closely aligned with the original data.


\subsection{Performance Comparison based on Captions Generated by Different LLMs}
\label{appendix:caption_llms}
We conduct a sensitivity study to examine whether the choice of large language models affects the generated dataset and the downstream performance.
We constructed a caption dataset containing positive and negative examples using three LLMs: Llama-3-8B \citep{dubey2024llama}, Qwen3-8B \citep{yang2025qwen3}, and GPT-4-nano \citep{achiam2023gpt4}.
Dataset construction yields 101,897 samples for the original pipeline, 98,261 samples with Qwen, and 101,852 samples with GPT.

All models use the same training configuration for fair comparison.
GPT-4-nano \citep{achiam2023gpt4} achieves the best performance on D${^3}$ with a score of $30.9$.
Performance remains consistent across datasets generated by different LLMs, which indicates low sensitivity to LLM choice.
Fig.~\ref{fig:vg_statistic} shows that the GPT-generated dataset exhibits a near-Gaussian word-length distribution.
The balanced caption lengths likely contribute to the superior performance.

\subsection{Human Evaluation for Generated Captions}
\label{appendix:caption_eval}
% \textcolor{blue}{We conducted a human evaluation with 50 participants on 500 randomly sampled queries from a 10 K dataset.
% Details related to the questionnaire are provided in Fig.~\ref{fig:survey}.
% Among the 500 evaluated samples, 6 were identified as mismatches. 
% The remaining 494 corresponded to the same object, resulting in a same-object rate of 98.8\%.
% The following are examples of failed cases.}

\begin{figure*}
    \centering
    \includegraphics[width=0.7\linewidth]{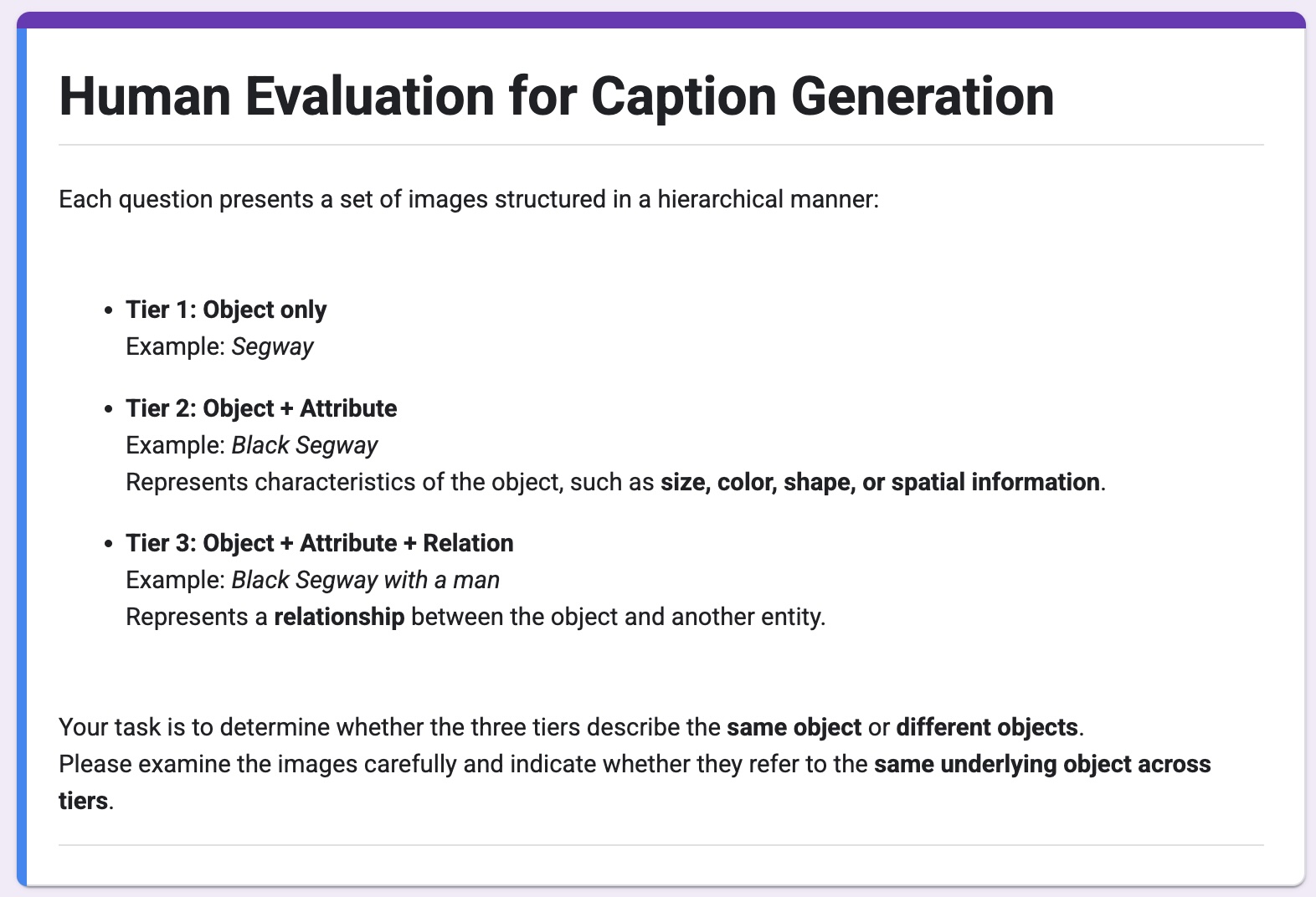}
    \vspace{0.1cm}
    \includegraphics[width=0.7\linewidth]{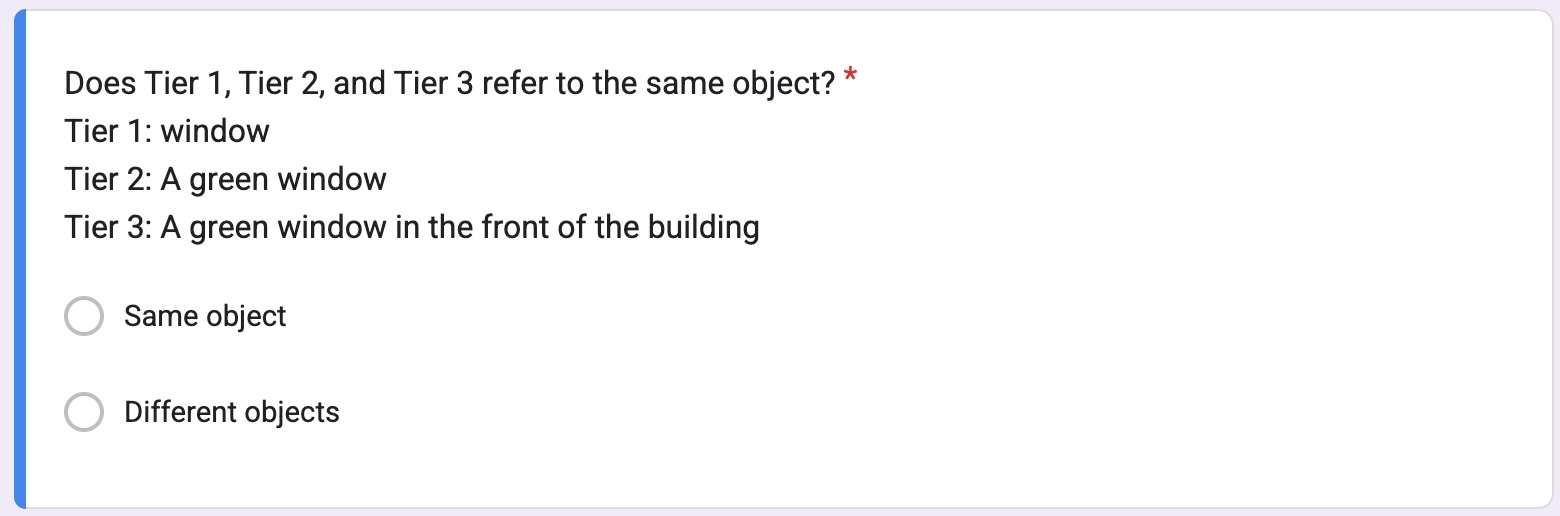}
    \vspace{0.1cm}
    \includegraphics[width=0.7\linewidth]{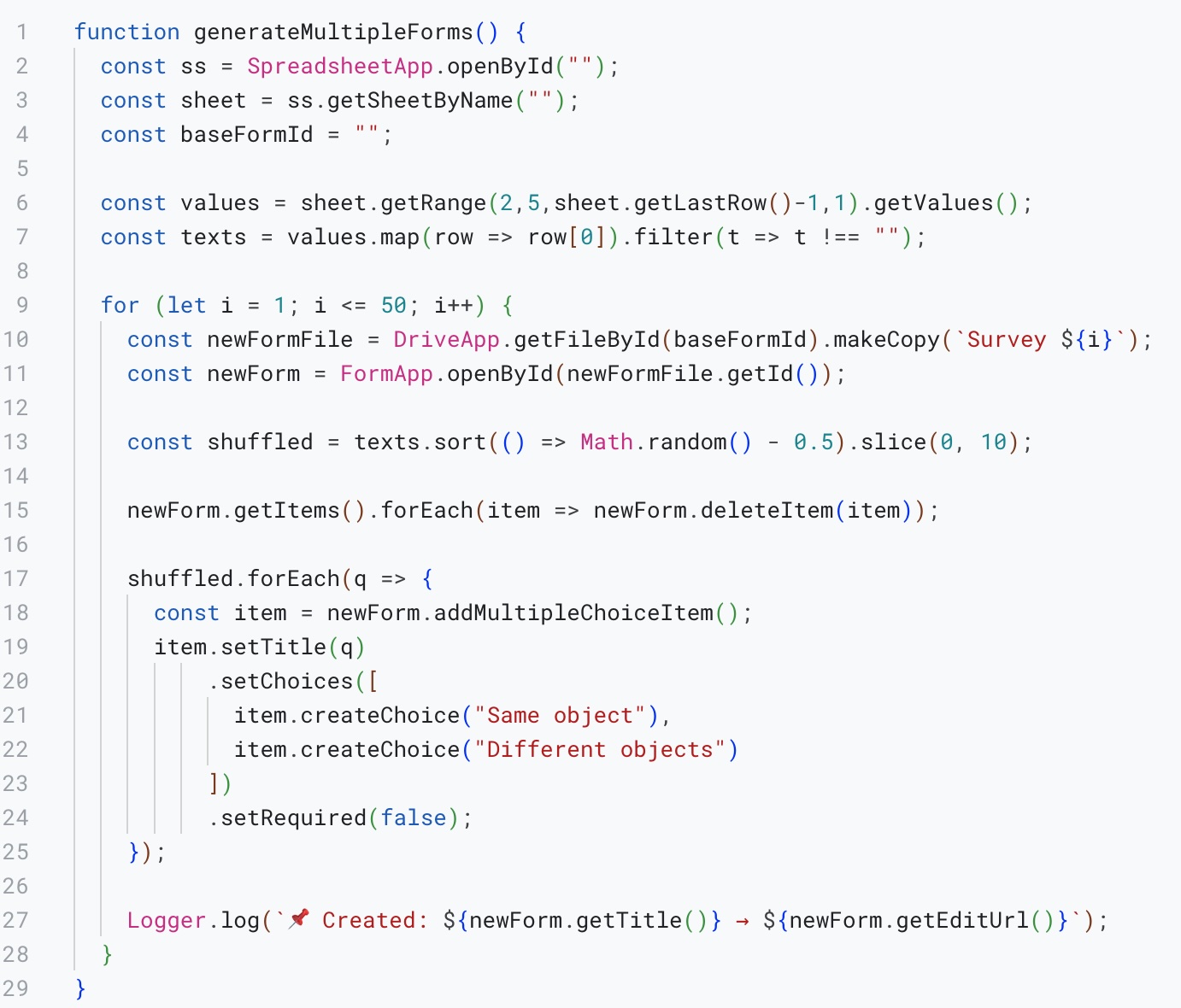}
    \caption{Google form-based questionnaire for human evaluation}
    \label{fig:survey}
\end{figure*}
To validate the quality of HiVG data, we conducted a human evaluation. Specifically, we aimed to assess whether the three-tier hierarchies in the LLM-generated captions accurately reflect our intended design. To this end, we asked 50 participants to verify whether the captions in each tier describe the same underlying objects. The caption sets were randomly assigned to participants, and a total of 500 samples were evaluated. A screenshot of the evaluation interface is provided in Fig.~\ref{fig:survey}.

The quality of the proposed HiVG dataset is remarkably high, supporting the effectiveness of our data-generation pipeline. 98.8\% of samples followed the intended hierarchy; only 6 of 500 depicted different target objects. The six failure cases are listed below:
\begin{tcolorbox}[
    colback=gray!10,
    colframe=gray!50,
    sharp corners,
    boxrule=0.7pt,
    left=4pt,
    right=4pt,
    top=4pt,
    bottom=4pt
]

\small
\begin{tabular}{l|p{2cm}|p{3.2cm}}
\textbf{Tier 1} & \textbf{Tier 2} & \textbf{Tier 3} \\ \hline
pen & a white pen & papers and a white pen on top of tab \\
cpu & a black cpu unit & a black cpu unit on the floor under a computer monitor \\
man & a black man & a bartender wearing a mustache and a tie \\
boat & the yellow boat & the yellow boat masts in the water \\
woman & woman and little girl & woman and little girl reading on a beige couch \\
numbers & red numbers & red numbers 2008 on the side of the building \\
\end{tabular}

\end{tcolorbox}

\section{Experimental Setup}
\label{sec:experimental_setting}

\textbf{Baselines}
\label{subsec:setup}
This paper compares the language-based object detection models on OFA~\citep{wang2022ofa}, OWL~\citep{owl}, G-DINO~\citep{gdino}, GLIP \citep{glip}, UNINEXT \citep{uninext}, Desco \citep{desco}, GLIP-GEN \citep{gen}, and GLEE \citep{glee}.

\subsection{Implementation Details}
We sample 16 images per batch and further select 6 corresponding sentences per image for hierarchy learning.
We employ AdamW~\citep{loshchilov2017decoupled} to optimize the trainable model, using a learning rate of $1\times10^{-4}$ for the \disenloss module and $5\times10^{-6}$ for LoRA.
For the comparison with baselines, our detector was trained for 60 K iterations, the same as in the ablation studies.
%LoRA was trained with a rank of 16 with a scaling factor of 16.
%We set the weights for the $\mathcal{L}_{class}$, $\mathcal{L}_{bbox}$, and $\mathcal{L}_{giou}$ to 4.0, 5.0, and 2.0, respectively. 
Following \citet{alper2024emergent}, the RE loss was set with a positive-to-negative ratio of 10:4.
In case of $\mathcal{L}_{\text{H}}$, we conduct experiments with positive-to-negative ratio of 2:1.
The value of $\gamma$ is set to 0.1.

\begin{table}[t]
\centering
    \begin{tabular}{c|c}
    \toprule
    Params.  & Value \\
    \midrule
    Batch size & 4 \\
    Optimizer & AdamW \\
    Optimizer momentum & $\beta_1=0.9$, $\beta_2 = 0.999$\\
    Rank of LoRA & 16 \\
    scaling factor of LoRA & 16 \\
    learning rate of LoRA & 5e-6 \\
    learning rate of \disenloss & 1e-4 \\
    Input resolution & 800 $\times$ 800 \\
    loss of class ($\mathcal{L}_{class}$) & 4.0 \\
    loss of bbox ($\mathcal{L}_{bbox}$) & 5.0 \\
    loss of gIoU ($\mathcal{L}_{giou}$) & 2.0 \\
    loss of TaSe ($\mathcal{L}_{\text{TaSe}}$) & 5.0 \\
    $\lambda$ & 0.1 \\
    \bottomrule
    \end{tabular}
    \caption{Hyperparameter settings}
    \label{tab:setting}
\end{table}

\subsection{Model Size and Budget}
For fine-tuning the pre-trained GLEE, we only train LoRA layers, \disenloss module, and VL fusion layers.
We train a total of 5,447,680 parameters, which is an efficient approach that reduces memory usage by 2.93\% of the model parameters.
On average, 10 iterations are executed, requiring 62.2 ms per inference. The disentangling and aggregation step adds an additional 1.57 ms (2.3\%) overhead.
Experiments were conducted using 4 NVIDIA A6000 GPUs for model training.
\begin{table}[t]
\centering
    \begin{tabular}{c|c}
    \toprule
    Layers  & \# of Params. (M) \\
    \midrule
    Image backbone & 23.5 \\
    Text encoder & 126.3 \\
    Detector & 31.5 \\
    LoRA & 0.4 \\
    \disenloss & 1.9 \\
    VL & 3.2 \\
    \midrule
    Trainable params. & 5.4 (2.93\%) \\
    \bottomrule
    \end{tabular}
    \caption{Parameter comparison across networks}
    \label{tab:configuration}
\end{table}

\subsection{Experimental Setup for Additional Detectors}
\label{sec:detectors_setup}
We apply TaSe to Grounding DINO-B \cite{gdino}, GLIP-T \cite{glip}, and Qwen3-VL-2B \cite{yang2025qwen3} to evaluate its model-agnostic effectiveness. We apply LoRA \cite{hu2021lora} and the disentanglement network, and tune only these modules. The learning rate is selected from $\{10^{-4}, 5\times10^{-5}, 10^{-5}\}$, with $5\times10^{-5}$ yielding the best validation performance across all models. The rank and alpha are fixed to 16. All models are trained only on the proposed HiVG data using the original training loss combined with $\mathcal{L}_{\text{TaSe}}$. We compute $\mathcal{L}_{\text{H}^{\pm}}$ using pooled embeddings to learn hierarchical structures.
For Qwen3-VL-2B, we use the corresponding tokenized span instead.

\noindent\textbf{GLIP Training Target Construction.}
Each class in GLIP-T needs a positive map constructed from the start and end indices. For the negative captions in HiVG, we randomly select one tier and concatenate the preceding and following strings to construct the positive map for training.

\noindent\textbf{Qwen3-VL Target Construction.}
We generate HiVG proposals and use their bounding boxes as prediction targets. We train the model with the cross-entropy loss and $\mathcal{L}_{\text{TaSe}}$.
%The training and inference prompts are provided below.

\section{Additional Results}

\subsection{Detailed Quantitative Results}
\label{appendix:detailed_qualitative_results}
\noindent\textbf{Detailed results on D$^3$ and OmniLabel.}
We report detailed results corresponding to the scores summarized in Tab. \ref{tab:results} and Tab. \ref{tab:grefcoco_odinw}. 
The average performance on the D3 and Omnilabel datasets, along with the results across different text lengths, is presented in Tab. \ref{tab:results_details}. 
Tab. \ref{tab:grefcoco} and Tab. \ref{tab:odinw} provide detailed results on the GRefCOCO \cite{he2023grec} and ODinW \cite{glip} datasets, respectively. 
The results consistently improve not only in terms of the average score but also across different text lengths.
To further analyze this trend, we conduct a more fine-grained evaluation by dividing the results into smaller text-length groups. 
As shown in the Tables, performance improvements are observed across multiple OOD datasets by adjusting only the text embeddings.

\begin{figure}[H]
    \includegraphics[
        width=\linewidth,
        trim=5 5 5 5,
        clip
    ]{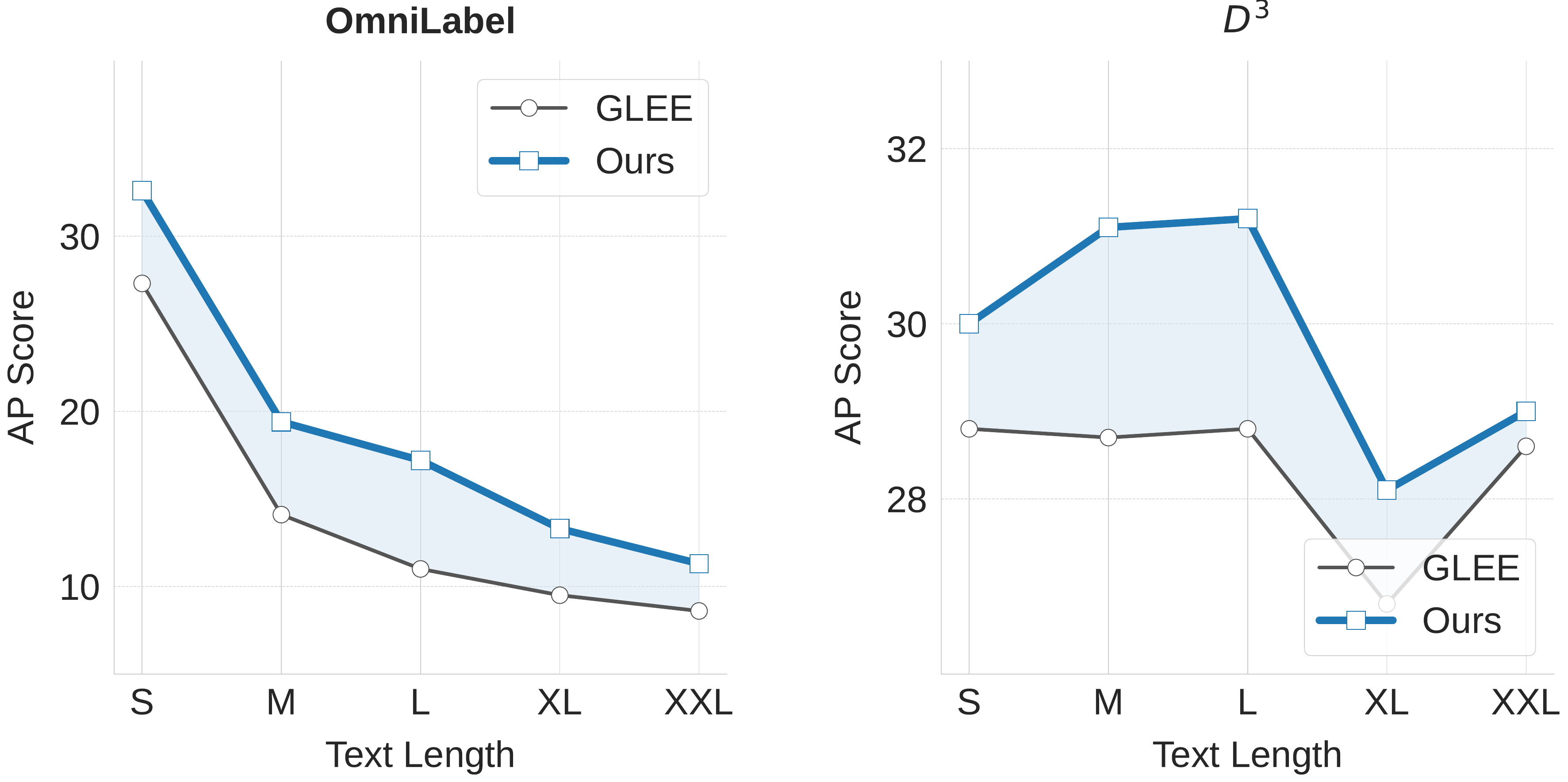}
    \caption{Performance analysis of Omnilabel and D$^3$ based on sentence length. S: [0, 3], M: [4, 7], L: [7, 10], XL: [10, 13], and XXL: [13, $\infty$].}
    \label{fig:text_length}
\end{figure}

\noindent\textbf{Performance comparison by text length.}
We further evaluate performance as a function of text length.
As illustrated in Fig. \ref{fig:text_length}, the proposed method achieves strong performance across all text-length ranges.
Decomposing into only three components (object, attribute, and relation) effectively models long textual descriptions for detection tasks and improves performance across all caption lengths.

\subsection{Additional Qualitative Results}
\begin{figure*}
    \centering
    \includegraphics[width=\linewidth]{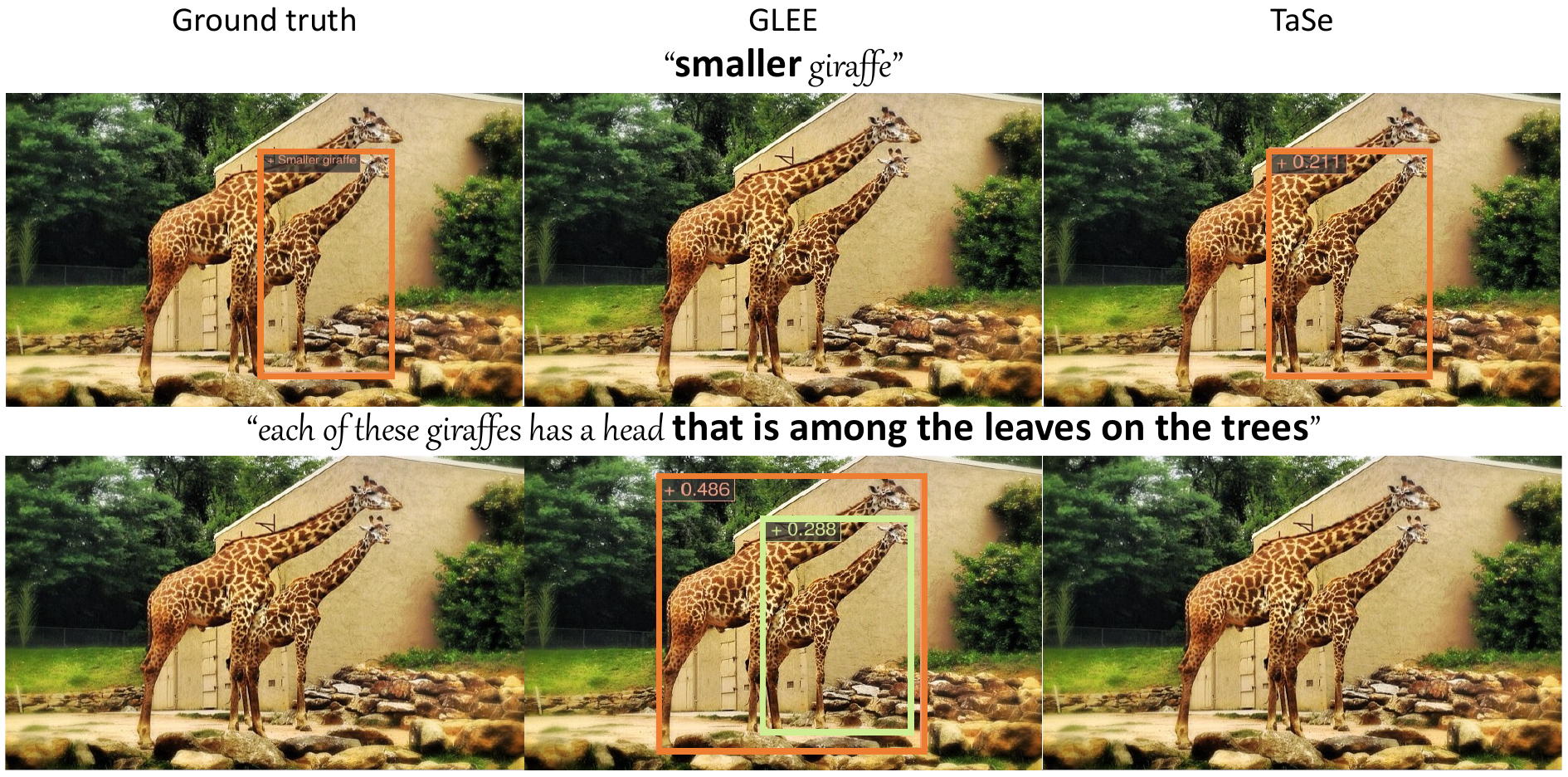}
    \includegraphics[width=\linewidth]{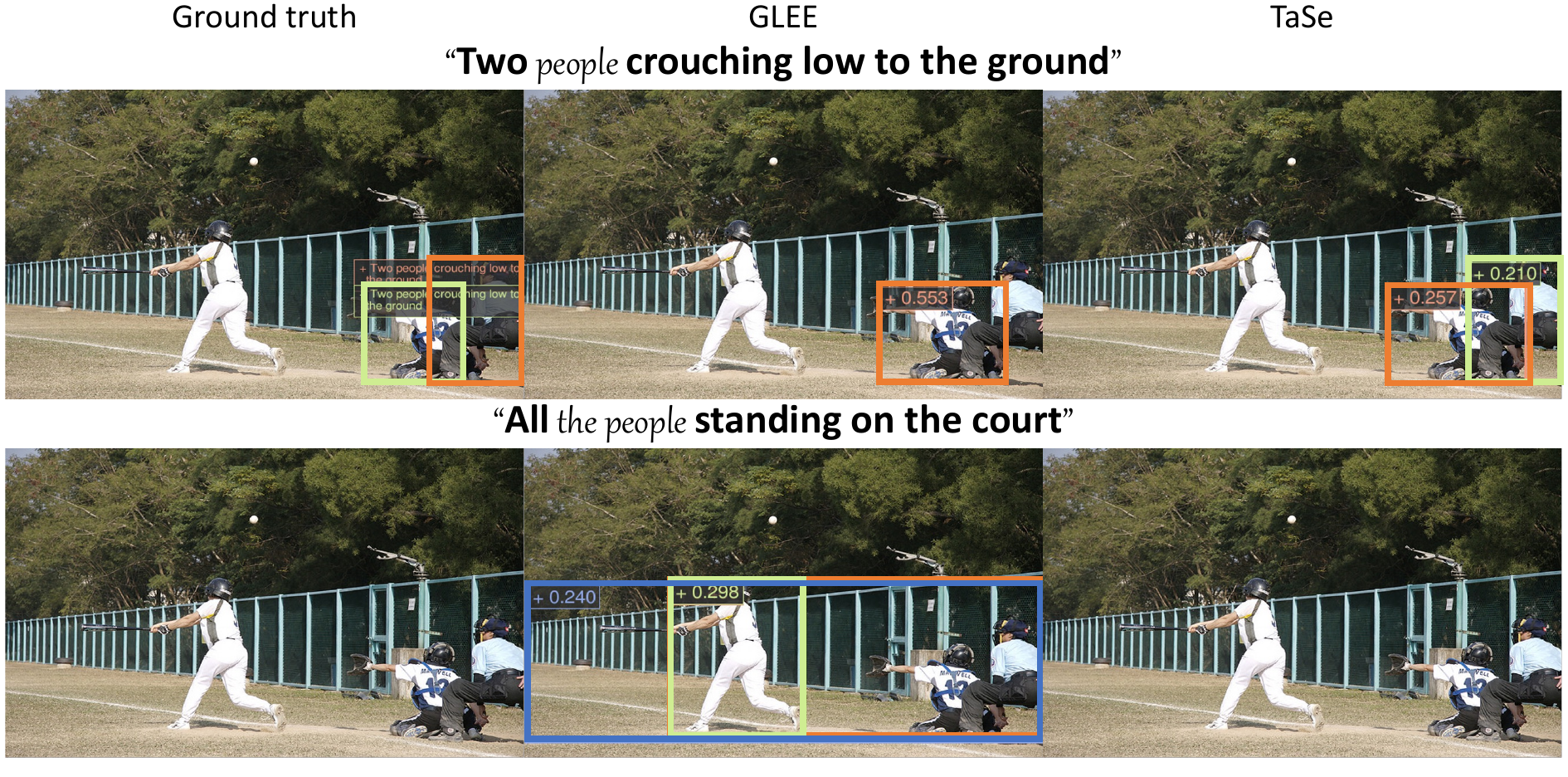}
    \caption{Qualitative analysis on Omnilabel data \cite{omnilabel}. We visualize and compare the results between GLEE and \model. We visualized the prediction results for both positive and negative captions of the same image.}
    \label{fig:quality_result2}
\end{figure*}

\begin{figure*}
    \centering
    \includegraphics[width=0.9\linewidth]{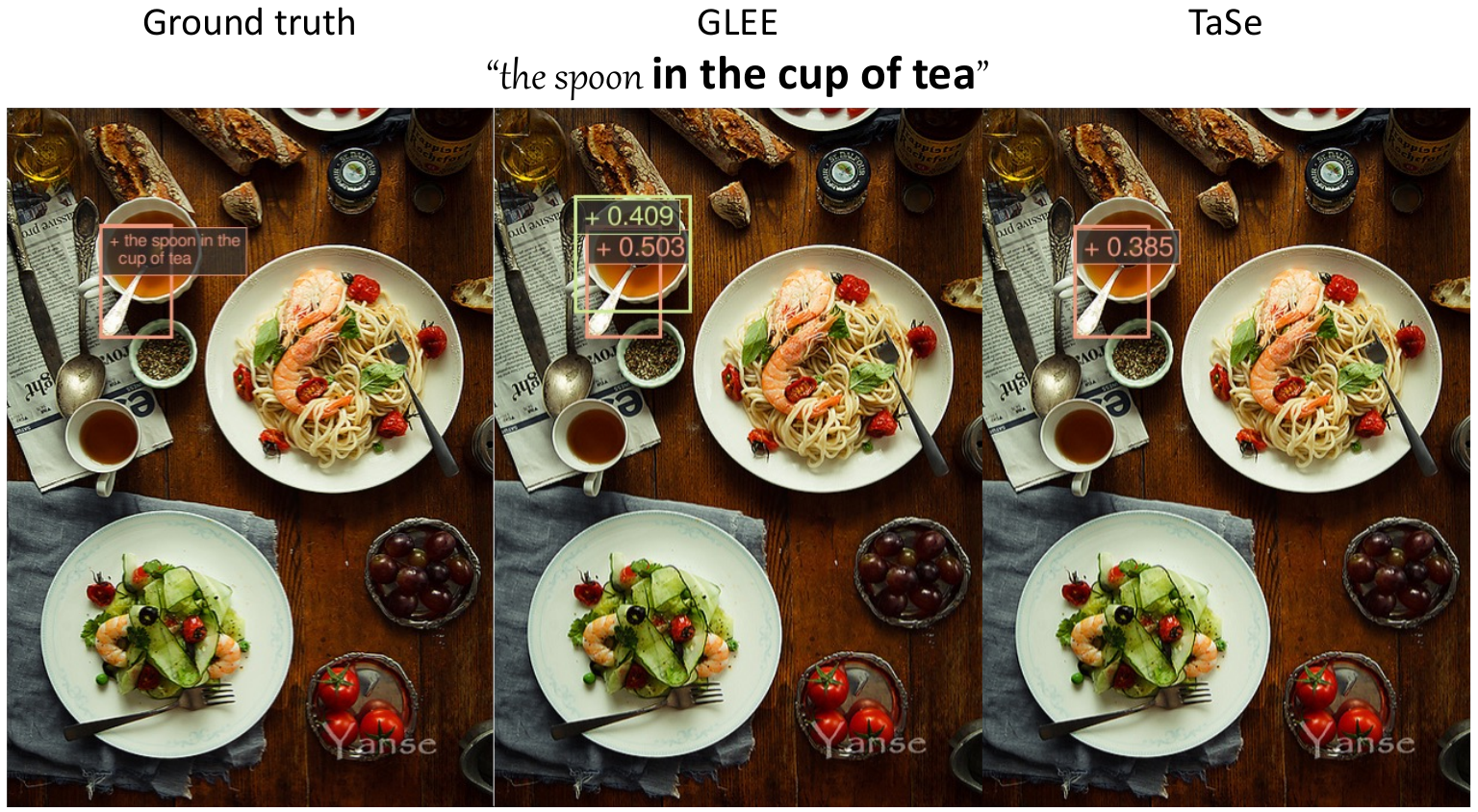}
    \includegraphics[width=0.9\linewidth]{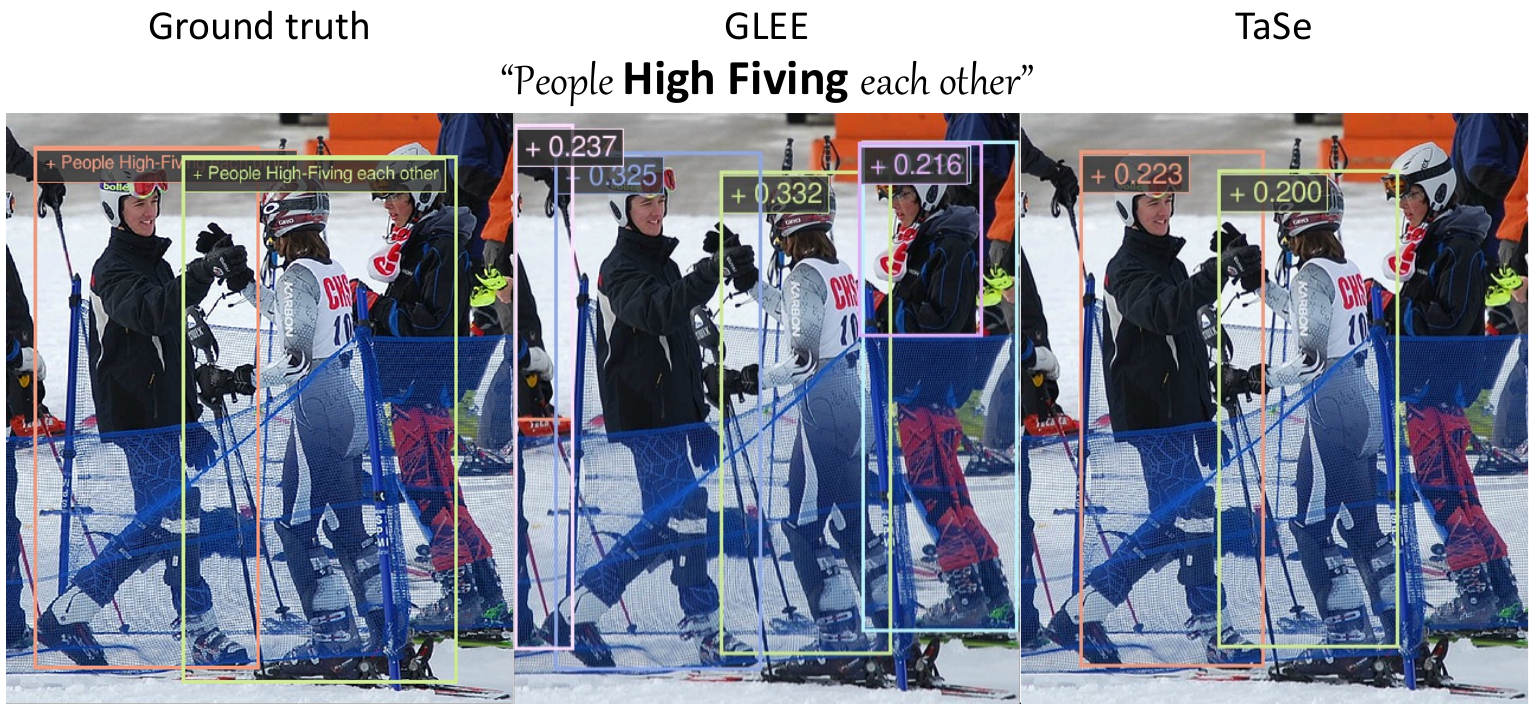}
    \caption{Qualitative analysis on Omnilabel data \cite{omnilabel}. We visualize and compare the results between GLEE and \model.}
    \label{fig:quality_result3}
    \centering
    \begin{minipage}{0.32\textwidth}
        \centering
        \includegraphics[width=\linewidth]{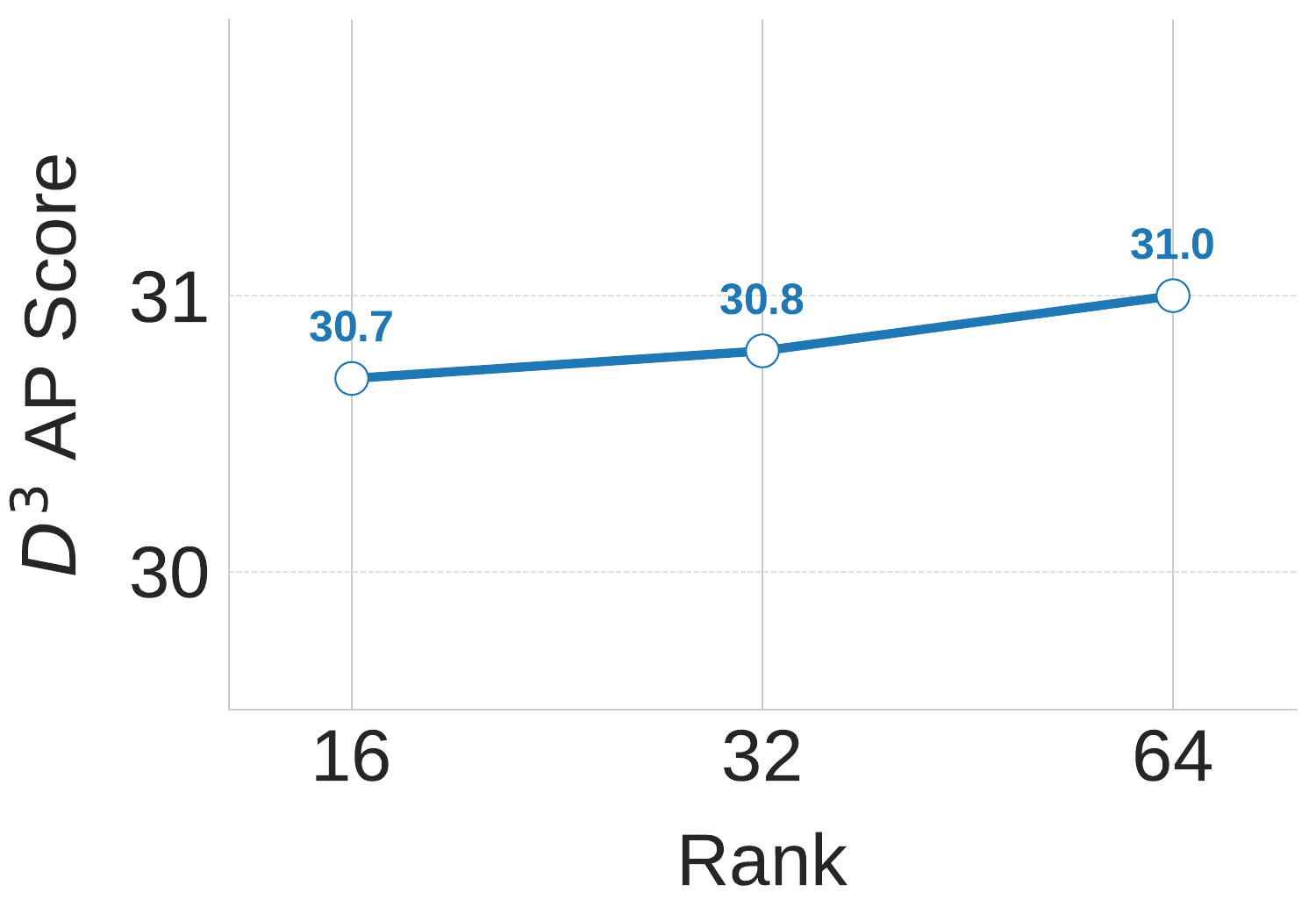}
        \caption{Analysis on the effect of different LoRA ranks}
        \label{fig:lora}
    \end{minipage}%
    \hfill
    \begin{minipage}{0.32\textwidth}
        \centering
        \includegraphics[width=\linewidth]{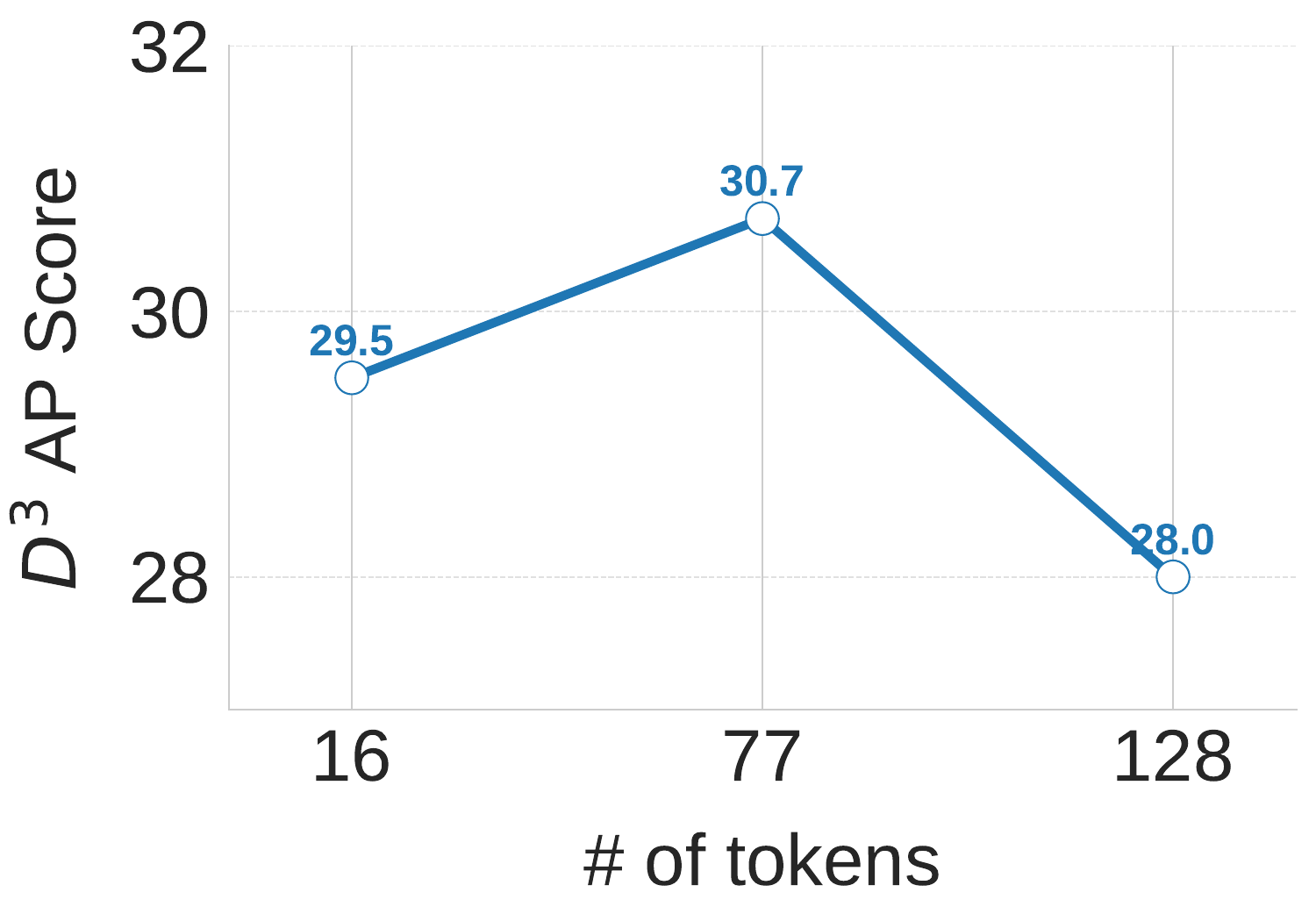}
        \caption{Analysis on the number of tokens}
        \label{fig:tokens}
    \end{minipage}
    \begin{minipage}{0.32\textwidth}
        \centering
        \includegraphics[width=\linewidth]{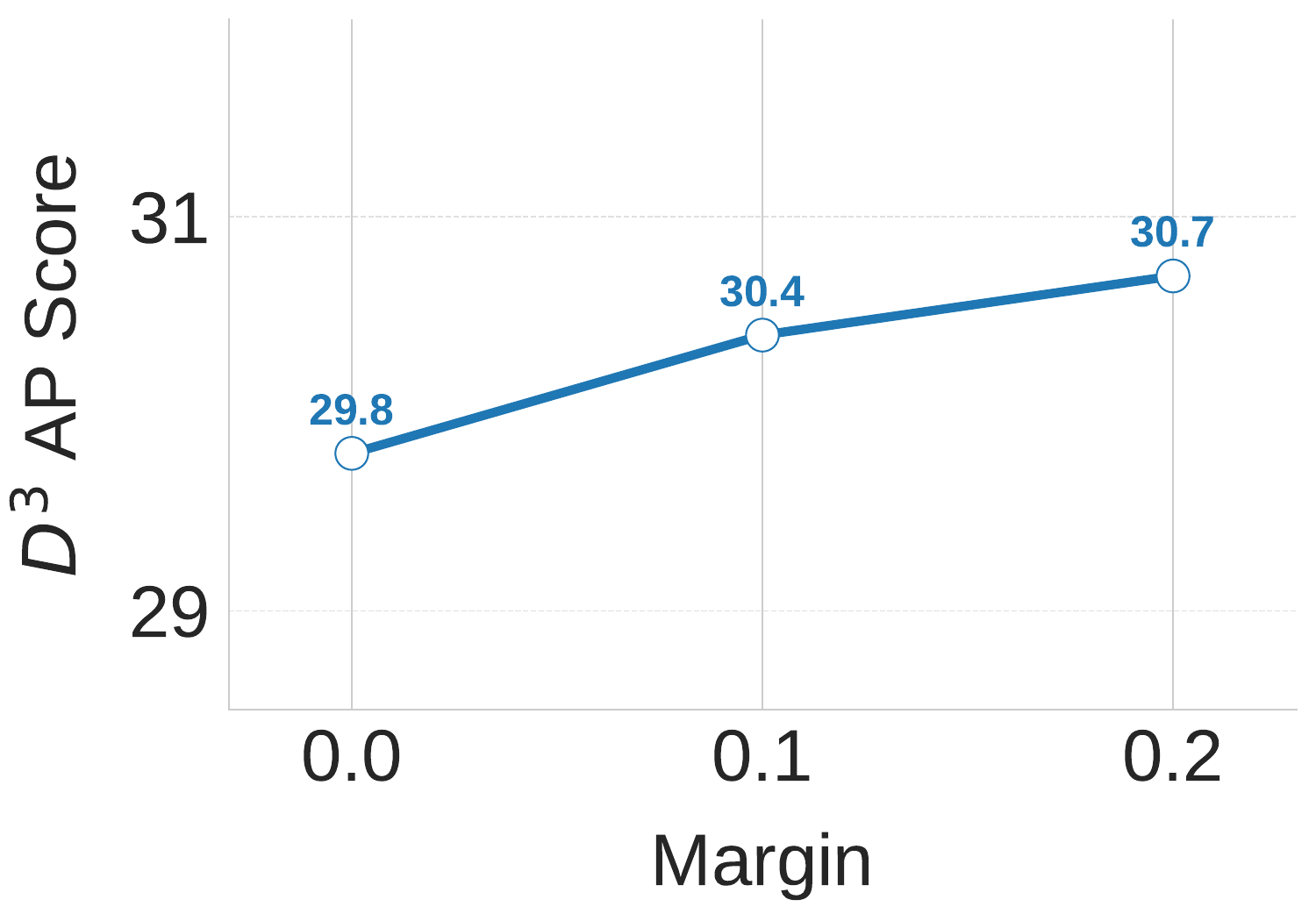}
        \caption{Analysis on the margin $m$ in the $\mathcal{L}_{\text{\disenloss}}$}
        \label{fig:margin}
    \end{minipage}
\end{figure*}

We present qualitative results to analyze the behavior of \model. 
For comparison, we evaluate text queries containing both attribute and relational cues. 
As shown in Fig. \ref{fig:quality_result2} and Fig. \ref{fig:quality_result3}, we visualize the results of two scenarios containing an absent example.
Given that sentences become longer, many VLMs focus on specific words, such as ``running" to detect objects.
For example, our model improves performance by capturing richer semantic information, such as the attribute ``pink,'' and by understanding contextual meaning, such as recognizing ``girl'' as the subject of ``running.''
In contrast, our model demonstrates greater robustness in detecting complex relations and predicts bounding boxes more accurately by better understanding object states and relative information.


%\section{Additional Experimental Results}

\subsection{Additional Ablation Studies}
\label{sec:ablation_details}
We conducted an ablation study on the model parameters. 
We compared and analyzed the effects of adjusting the LoRA \citep{hu2021lora} rank (see Fig.~\ref{fig:lora}), the number of tokens (see Fig.~\ref{fig:tokens}), and the margin parameter (see Fig.~\ref{fig:margin}) for the disentangled loss.
Performance remains stable across different parameter settings.
Positive and negative sentences in higher tiers partially share the embedding space due to hierarchical semantic structures.
Strict disentanglement constrains representation learning.
Token disentanglement and aggregation benefit from preserving the original token length.
Maintaining the standard sequence length of 77 tokens preserves positional information and stabilizes token representations during manipulation.
Increasing the LoRA rank yields moderate performance improvements.
Rank 16 provides the best balance between performance and parameter efficiency.

\setlength{\tabcolsep}{2.0pt}
\begin{table*}[t]
\centering
\resizebox{\textwidth}{!}{
\begin{tabular}{lc *{13}{>{\centering\arraybackslash}p{1.1cm}}}
\toprule
\multirow{2}{*}{\textbf{Model}} 
& \multirow{2}{*}{\textbf{Param.}} 
& \multicolumn{3}{c}{\textbf{D$^3$ (default)}} 
& \multicolumn{4}{c}{\textbf{D$^3$ (length)}} 
& \multicolumn{3}{c}{\textbf{OmniLabel (default)}} 
& \multicolumn{3}{c}{\textbf{OmniLabel (length)}} \\
\cmidrule(r){3-5} 
\cmidrule(lr){6-9} 
\cmidrule(lr){10-12} 
\cmidrule(lr){13-15}
& 
& \textbf{Full} & \textbf{Pres} & \textbf{Abs} 
& \textbf{S} & \textbf{M} & \textbf{L} & \textbf{XL} 
& \textbf{AP} & \textbf{AP$_c$} & \textbf{AP$_d$} 
& \textbf{S} & \textbf{M} & \textbf{L} \\
\midrule

\multicolumn{15}{l}{Specialist models} \\

OFA-L~\citep{wang2022ofa} 
& 0.5B 
& 4.2 & 4.1 & 4.6 
& 4.9 & 5.4 & 3.0 & 2.1 
& 2.7 & 2.7 & 2.6 
& 3.6 & 2.7 & 2.3 \\

MDETR~\citep{mdetr} 
& 0.1B 
& - & - & - 
& - & - & - & - 
& - & 4.7 & 9.1 
& 6.4 & 4.6 & 4.0 \\

OWL~\citep{owl} 
& 0.2B 
& 9.6 & 10.7 & 6.4 
& 20.7 & 9.4 & 6.0 & 5.3 
& 8.0 & 15.6 & 5.4 
& 5.7 & 5.4 & 6.2 \\

UNINEXT~\citep{uninext} 
& - 
& 21.6 & 23.7 & 15.4 
& 23.6 & 22.6 & 20.5 & 18.4 
& 22.2 & 27.2 & 18.8 
& - & - & - \\

\rowcolor{gray!15} 
G-DINO-B~\citep{gdino} 
& 0.2B 
& 20.7 & 20.1 & 22.5 
& 22.6 & 22.5 & 18.9 & 16.5 
& 19.3 & 23.6 & 16.4 
& 29.4 & 14.8 & 8.2 \\

GEN~\citep{gen} 
& - 
& 21.4 & 20.6 & 23.7 
& 28.1 & 24.5 & 17.4 & 11.5 
& 22.2 & 27.2 & 18.8 
& - & - & - \\

\rowcolor{gray!15} 
GLIP-T~\citep{glip} 
& 0.2B
& 19.1 & 18.3 & 21.5 
& 22.4 & 22.0 & 16.6 & 10.6 
& 19.3 & 23.6 & 16.4 
& 29.4 & 14.8 & 8.2 \\

GLIP-T + \desco~\citep{desco} 
& 0.2B
& 24.2 & 22.9 & 27.8 
& 24.3 & 21.9 & 16.4 & 11.5 
& 23.8 & 27.4 & 21.0 
& 33.7 & 19.0 & 13.7 \\

\rowcolor{gray!15} 
GLEE-Lite$^{*}$~\citep{glee} 
& 0.2B 
& 27.6 & 26.8 & 30.1 
& 30.0 & 27.6 & 26.9 & 17.2 
& 21.7 & 36.6 & 15.4 
& 28.4 & 13.8 & 10.3 \\

\midrule
\multicolumn{15}{l}{Generalist models} \\

\rowcolor{gray!15} 
Qwen3-VL-2B~\citep{yang2025qwen3} 
& 2B 
& 14.9 & 14.5 & 16.1 
& 11.0 & 18.1 & 13.8 & 11.5 
& - & - & 3.9 
& 3.3 & 4.6 & 3.9 \\

InternVL2-8B~\citep{internvl} 
& 8B 
& 9.8 & 11.0 & 6.2 
& - & - & - & - 
& - & - & 4.5 
& 6.2 & 4.2 & 3.9 \\

Griffon~\citep{griffon} 
& 13B 
& 12.3 & 12.4 & 12.2 
& - & - & - & - 
& - & - & 7.2 
& 12.5 & 6.3 & 4.5 \\

Groma~\citep{groma} 
& 7B 
& 16.0 & 15.9 & 16.3 
& - & - & - & - 
& - & - & 15.6 
& 21.4 & 15.6 & 11.3 \\

ROD-MLLM~\citep{rodmllm} 
& 7B 
& 29.7 & 30.0 & 28.7 
& - & - & - & - 
& - & - & 25.3
& 31.8 & 24.5 & 21.0 \\

GPT-5.4-nano~\cite{gpt5} 
& - 
& 1.0 & 3.5 & 3.5 
& 1.5 & 1.2 & 0.1 & 0.1 
& - & - & 3.9 
& 3.3 & 4.6 & 3.9 \\

\midrule
\multicolumn{15}{l}{Ours} \\

\rowcolor{yellow!20} 
GLEE-Lite + TaSe 
& 0.2B 
& 30.7 & 29.9 & 33.2
& 31.8 & 31.2 & 30.3 & 19.8
& 26.9 & 36.8 & 21.2 
& 33.1 & 19.3 & 14.8 \\

\rowcolor{yellow!20} 
G-DINO-B + TaSe 
& 0.2B 
& 24.3 & 22.9 & 28.5 
& 27.7 & 26.2 & 22.4 & 18.8
& 28.6 & 40.5 & 22.1 
& 31.1 & 44.7 & 18.7 \\

\rowcolor{yellow!20} 
GLIP-T + TaSe 
& 0.2B 
& 21.5 & 21.2 & 22.7 
& 26.4 & 23.3 & 18.8 & 16.0 
& 21.5 & 24.3 & 19.9 
& 34.6 & 17.3 & 10.5 \\

\rowcolor{yellow!20} 
Qwen3-VL-2B + TaSe 
& 2B 
& 41.6 & 43.0 & 37.3 
& 42.7 & 44.1 & 38.3 & 38.7
& - & - & 30.4 & 40.3 & 29.1 & 25.1 \\

\bottomrule
\end{tabular}}
\vspace{-2mm}
\caption{
Evaluation on D$^3$~\citep{d3} and OmniLabel~\citep{omnilabel}. 
D$^{3}$ provides three types of descriptions: absence (ABS), presence (PRES), and full (FULL). 
Models are grouped into specialist models, generalist models, and ours. 
Param. indicates the publicly reported model size when available. 
For MLLM-based baselines, only AP$_d$ is reported on OmniLabel. 
For OmniLabel, the final AP is computed as the geometric mean of category-level (AP$_c$) and description-level (AP$_d$) scores. 
}
\label{tab:results_details}
\end{table*}%
\begin{table*}[t!]
  \renewcommand\arraystretch{1.05}
  \small
  \centering
     \setlength{\tabcolsep}{0.96mm}{\begin{tabular}{l|c|c|cc|cc|cc}
      \specialrule{.1em}{.05em}{.05em} 
      \multirow{2}{*}{Methods} &\multirow{2}{*}{\shortstack{Visual\\Encoder}} & \multirow{2}{*}{\shortstack{Textual\\Encoder}}& \multicolumn{2}{c|}{val} & \multicolumn{2}{c|}{testA} & \multicolumn{2}{c}{testB} \\
             &  &  &Pr@0.5   & N-acc.   &Pr@0.5  &  N-acc.  &Pr@0.5   & N-acc. \\
        \hline\hline
        MCN~\cite{luo2020multi} & DarkNet-53 &GRU& 28.0 &30.6  & 32.3 & 32.0 & 26.8 & 30.3 \\
        VLT~\cite{ding2021vision}  & DarkNet-53 & GRU& 36.6 & 35.2  & 40.2 & 34.1 & 30.2 & 32.5 \\
        MDETR & ResNet-101 & RoBERTa& 42.7 & 36.3 & 50.0 & 34.5 & 36.5 & 31.0   \\
        UNINEXT$^\dagger$~\cite{uninext} & ResNet-50 & BERT&58.2  &  50.6 & 46.4 & 49.3 & 42.9 & 48.2 \\
        \rowcolor{gray!15} GLEE \citep{glee} & RN50 & CLIP & 66.7 & 62.4 & 73.0 & 64.1 & 61.8 & 51.9 \\
        \rowcolor{yellow!20} GLEE + TaSe & RN50 & CLIP & \textbf{69.6} & \textbf{69.9} & \textbf{76.6} & \textbf{69.6} & \textbf{67.7} & \textbf{58.9} \\
        \specialrule{.1em}{.05em}{.05em} 
    \end{tabular}}
\vspace{-2mm}
\caption{Zero-shot evaluation results on gRefCOCO \citep{he2023grec}. Following \cite{he2023grec}, Pr@0.5 (F1=1, IoU$\geq$0.5) denotes a true positive when the predicted bounding box achieves an IoU greater than 0.5 with the ground truth. N-acc. represents the positive rate (TN / Positive).}
\label{tab:grefcoco}
\end{table*}%
\begin{table*}[t!]
\small
\newcolumntype{C}{>{\centering\arraybackslash}p{0.7cm}}
\centering
\resizebox{0.85\linewidth}{!}{
\begin{tabular}{@{} l|CCCCCCCCCCCCC|C}
\toprule
& \rotatebox{90}{Pascal VOC}
& \rotatebox{90}{Aerial Drone}
& \rotatebox{90}{Aquarium}
& \rotatebox{90}{Rabbits}
& \rotatebox{90}{Ego Hands}
& \rotatebox{90}{Mushrooms}
& \rotatebox{90}{Packages}
& \rotatebox{90}{Raccoon}
& \rotatebox{90}{Shellfish}
& \rotatebox{90}{Vehicles}
& \rotatebox{90}{Pistols}
& \rotatebox{90}{Pothole}
& \rotatebox{90}{Thermal}
& \rotatebox{90}{Average} \\
\midrule
\rowcolor{gray!15} GLEE & 61.2 & 5.0 & 23.9 & 71.9 & 46.2 & \textbf{57.8} & 25.6 & \textbf{56.8} & 33.1 & 60.6 & 57.1 & \textbf{25.3} & \textbf{52.5} & 44.4 \\
\rowcolor{yellow!20} 
\model & \textbf{61.3} & \textbf{14.4} & \textbf{24.8} & \textbf{81.4} & \textbf{54.5} & 20.8 & \textbf{63.1} & 50.6 & \textbf{41.0} & \textbf{60.8} & \textbf{61.5} & 18.9 & 46.7 & \textbf{49.3} \\
\bottomrule
\end{tabular}
}
\vspace{-2mm}
\caption{Zero-shot evaluation results across 13 datasets in ODinW~\citep{li2022elevater}.}
\label{tab:odinw}
\end{table*}

\subsection{Pseudo Code for TaSe}
We provide the details of the disentanglement modes employed for compositional learning as shown in Alg.~\ref{alg:compositional_learning}.
The first mode adopts traditional mean-pooling and uses the resulting representation for contrastive learning.
The second and third modes involve disentanglement via a \disenloss module, followed by contrastive learning based on the aggregated compositional embedding.
Specifically, the second mode applies the \disenloss module at the token level to leverage information across all tokens, whereas the third mode applies the module after pooling, focusing on sentence-level semantics.
{\footnotesize
\begin{algorithm}[!t]
\caption{Pseudo code for disentangling and hierarchical aggregation paradigm}
\label{alg:compositional_learning}
\setlength{\algomargin}{8pt}
\SetAlgoNlRelativeSize{-1}

\KwIn{
Image-text pair $\left(\text{v}_i, t_i\right)_{i=1}^{B}$; 
Text encoder with LoRA $\mathcal{T}_{\theta}$; 
Learnable vectors $\mathbf{V}_{\text{O}}, \mathbf{V}_{\text{A}}, \mathbf{V}_{\text{R}}$; 
Projection embedding $\delta$; 
Vision backbone $\mathcal{V}_{\psi}$; 
MaskDINO $f_{\psi}$}

\KwOut{Bounding box and class $Y$; Total Loss $\mathcal{L}$;}

\For{each training iteration}{
$\mathbf{X}_{\text{v}} = \mathcal{V}_{\psi}(\text{v})$\\
$\mathbf{X} = \mathcal{T}_{\theta}(t) \cdot \delta$\\
\tcp{\disenloss}
$\mathbf{X} = \mathbf{FFN(X)}$\\
$\mathbf{O} =$ CrossAttn($\mathbf{X, V_O}$)\\
$\mathbf{A} =$ CrossAttn($\mathbf{X, V_A}$)\\
$\mathbf{R} =$ CrossAttn($\mathbf{X, V_R}$)\\
$\mathbf{E} = \text{Pool}(\text{FFN}(\mathbf{O+A+R}))$\\
$\mathbf{Y}, \mathcal{L} = f_{\psi}(\mathbf{X}_{\text{v}}, \mathbf{E})$
}
Update $\theta$ by minimizing $\mathcal{L}$
\end{algorithm}
}
\section{Discussions}

\subsection{How Are Embeddings Structured After Disentanglement?}
While a few negative pairs still exhibit large angles, Fig.~\ref{fig:angle} confirms that positive and negative pairs are effectively aligned in the embedding space. 
To examine whether this alignment follows the intended structure after disentanglement and hierarchical aggregation, we visualize the t-SNE projection of the trained \model embeddings.
As shown in Fig.~\ref{fig:emb_glee}, GLEE is dispersed around category names, whereas \model realigns embeddings around objects and preserves robust angular distance for negatives corresponding to attributes or relations. This is evident in the ``segway'' object, where the captions ``black segway with a woman'' and ``black segway with a man'' lie at different angles from the reference point ``black segway.''

\subsection{Comparative Analysis of Text Embeddings across State-of-the-Art Vision-Language Models}
VLMs are pre-trained for text–image alignment, and Fig.~\ref{fig:fig1_2} shows that they often demonstrate bag-of-words behavior.
Recent VLMs incorporate approaches such as MAE-based architectures \citep{fang2024eva} and LLM-connected visual encoders \citep{jiang2024e5} to improve representation quality. 
We investigate whether the text embeddings of these models improve object detection. For this analysis, we visualize text embeddings from SigLIP \citep{zhai2023sigmoid} and E5-V \citep{jiang2024e5} using the same example shown in Fig.~\ref{fig:siglip}.

Fig.~\ref{fig:siglip} shows that SigLIP demonstrates embeddings that closely align across distinct objects (e.g., black segway with a man and black segway with a woman), similar to CLIP. For E5-V, the LLM-based embeddings were more object-centered (e.g., woman), yet white segway and black segway remained closer to each other than black segway with a man. 
This observation indicates that even advanced VLMs can retain ambiguities in object-level representation, as they rely on textual embedding similarity to correlate visual content.

\subsection{Exterior Angle Analysis}
\label{sec:exterior_angle_analysis}
We validate the effectiveness of our proposed hierarchical learning approach by visualizing the impact of the angular loss.
For the experimental setup, we randomly initialize 50 two-dimensional embeddings and train them using the original hierarchy loss and our extended loss function.

Based on the results, the baseline method pushes both positive and negative samples away in the embedding space. 
The proposed method pulls positive samples closer and pushes negative samples farther apart. 
The proposed method moves text embeddings with stable optimization and limited training.

\begin{figure}
    \centering
    \includegraphics[width=0.95\linewidth, trim={0 0.7cm 0 0}, clip]{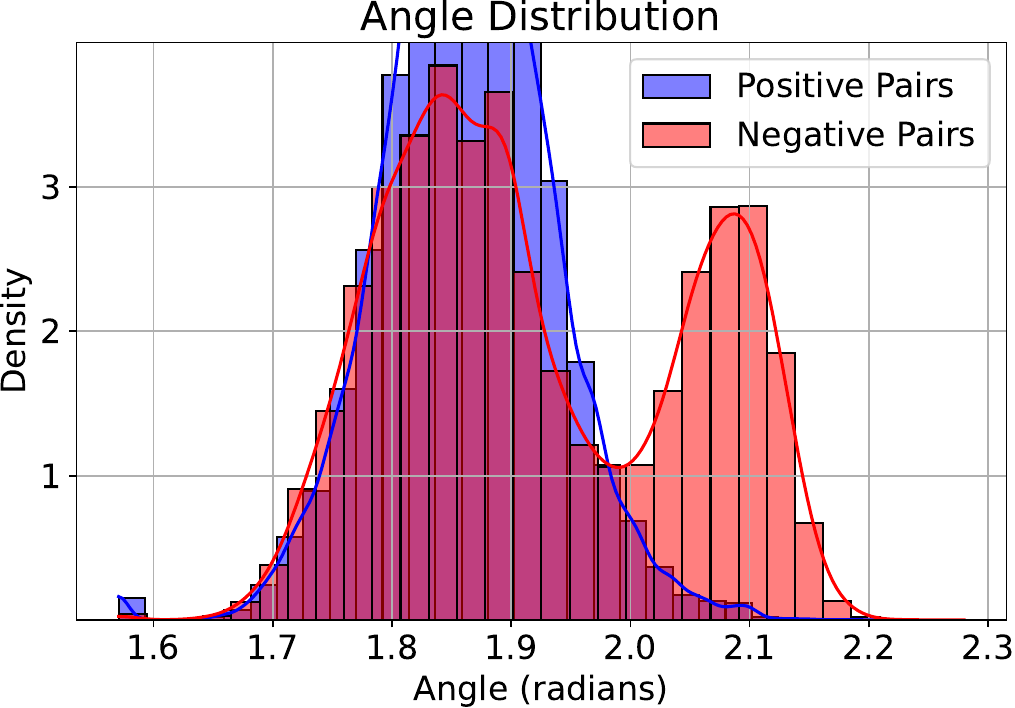}
    \includegraphics[width=0.95\linewidth, trim={0 0 0 0.65cm}, clip]{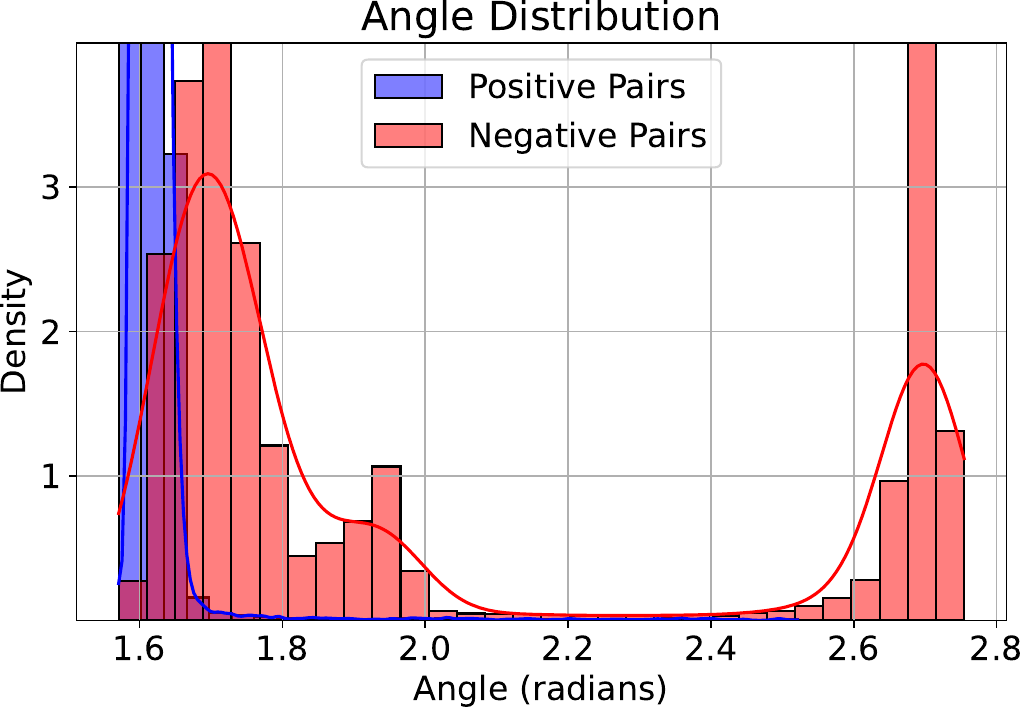}
    \vspace{-1em}
    \caption{Comparison of exterior angles between GLEE and \model}
    \label{fig:angle}
\end{figure}

\begin{figure*}[t]
\centering
\vspace{-4mm}

% Row 1
\begin{subfigure}{0.32\linewidth}
\includegraphics[width=\linewidth]{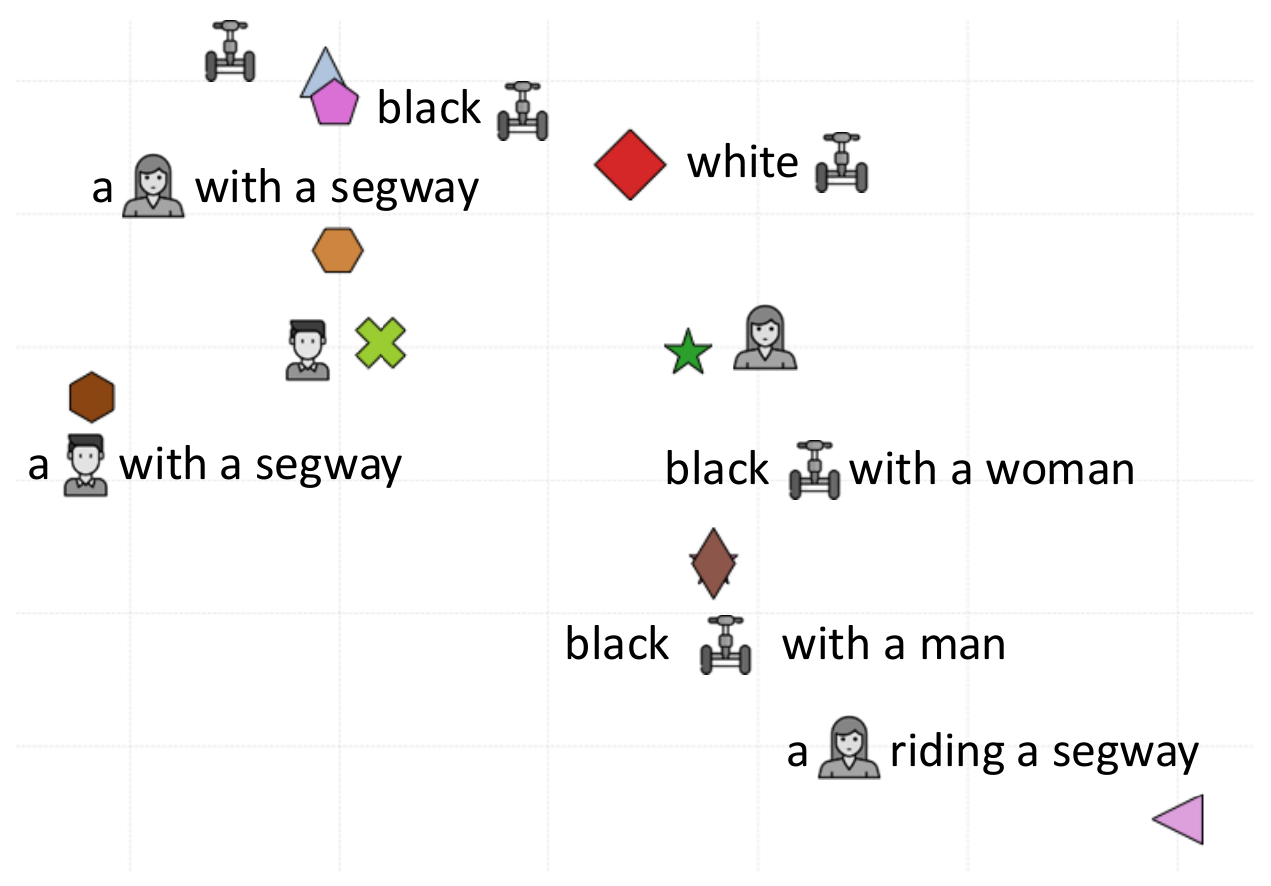}
\caption{GLEE embedding visualization}
\label{fig:emb_glee}
\end{subfigure}
\hfill
\begin{subfigure}{0.32\linewidth}
\includegraphics[width=\linewidth]{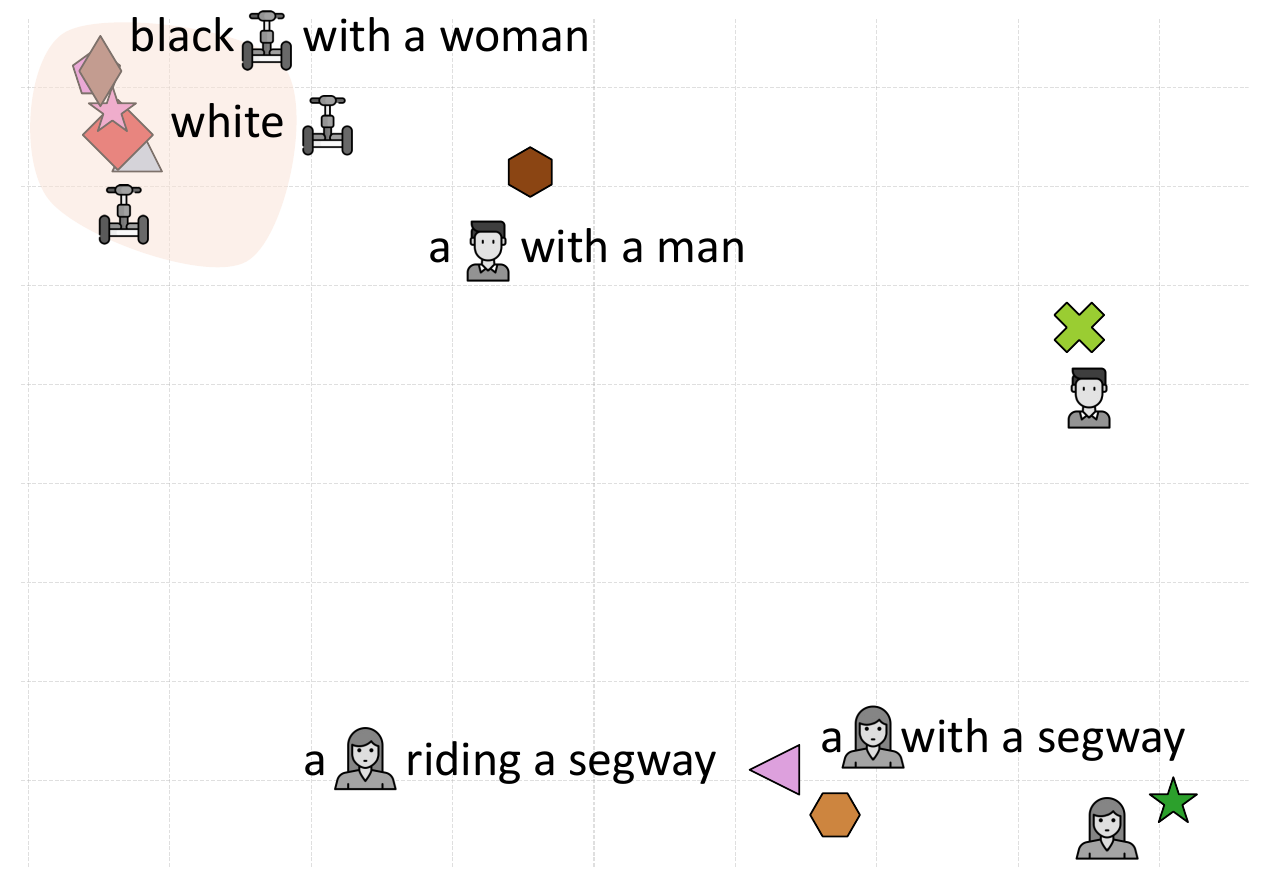}
\caption{\model embedding visualization}
\end{subfigure}
\hfill
\begin{subfigure}{0.32\linewidth}
\includegraphics[width=\linewidth]{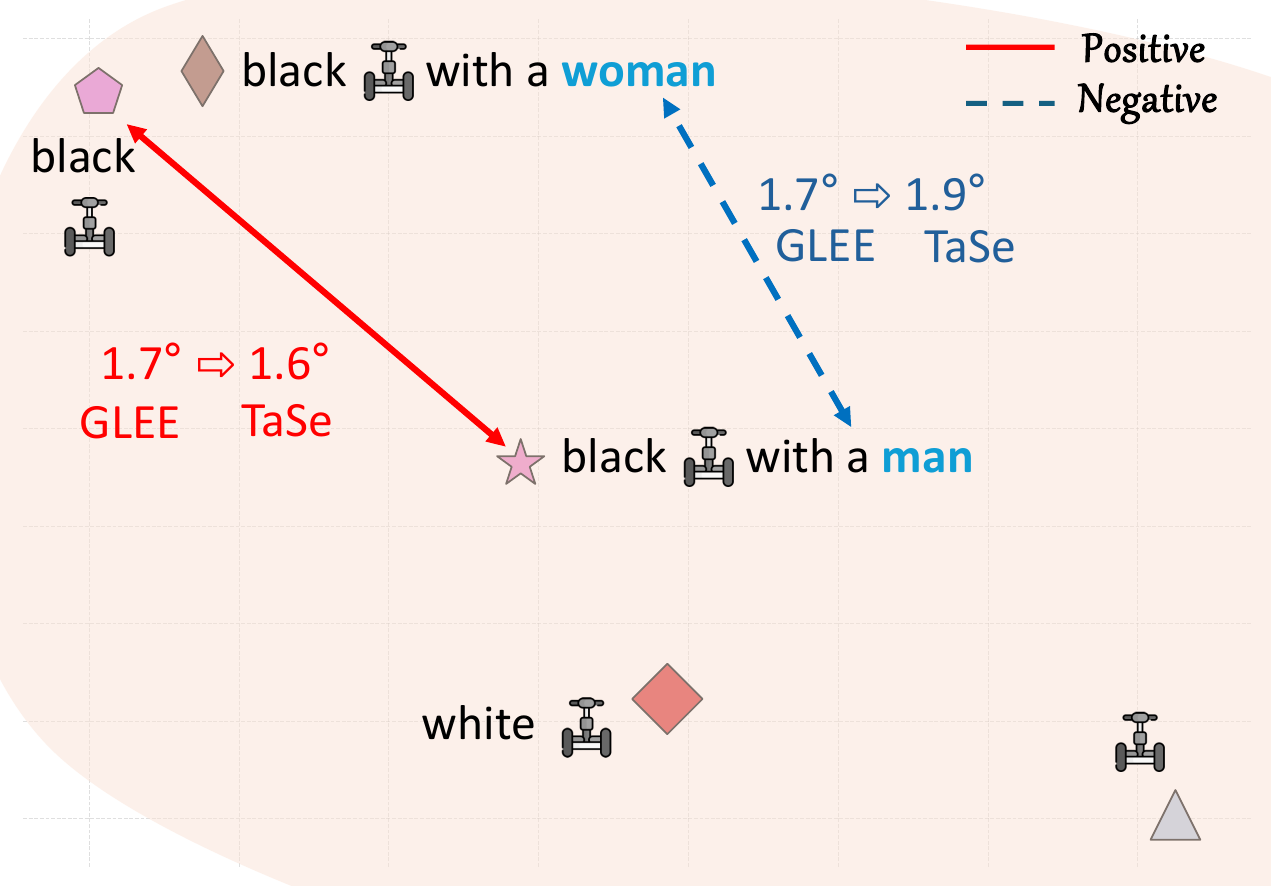}
\caption{Zoom in \model embedding}
\end{subfigure}

\vspace{2mm}

% Row 2
\begin{subfigure}{0.32\linewidth}
\centering
\includegraphics[width=\linewidth]{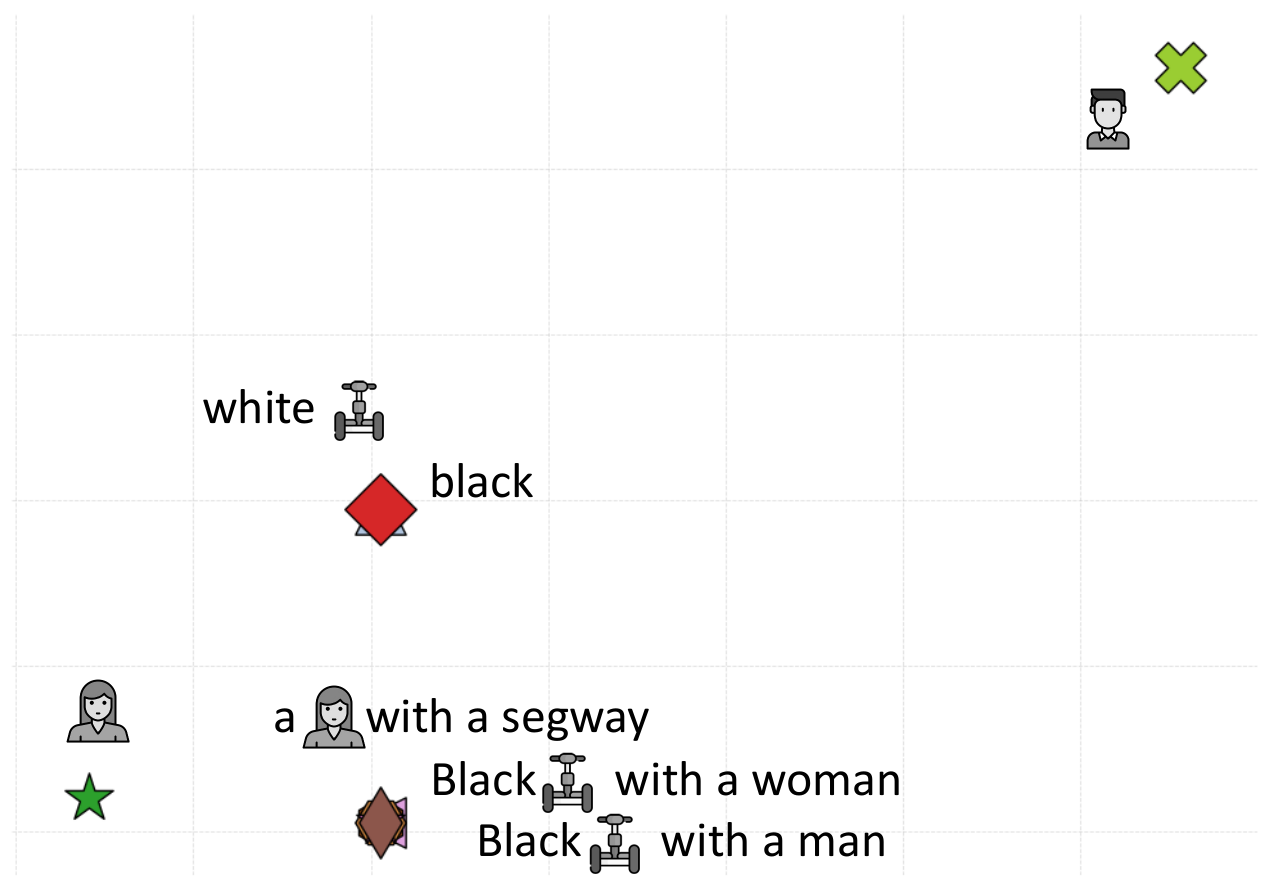}
\caption{SigLIP text embedding}
\label{fig:siglip}
\end{subfigure}
\vspace{1.5em}
\begin{subfigure}{0.32\linewidth}
\centering
\includegraphics[width=\linewidth]{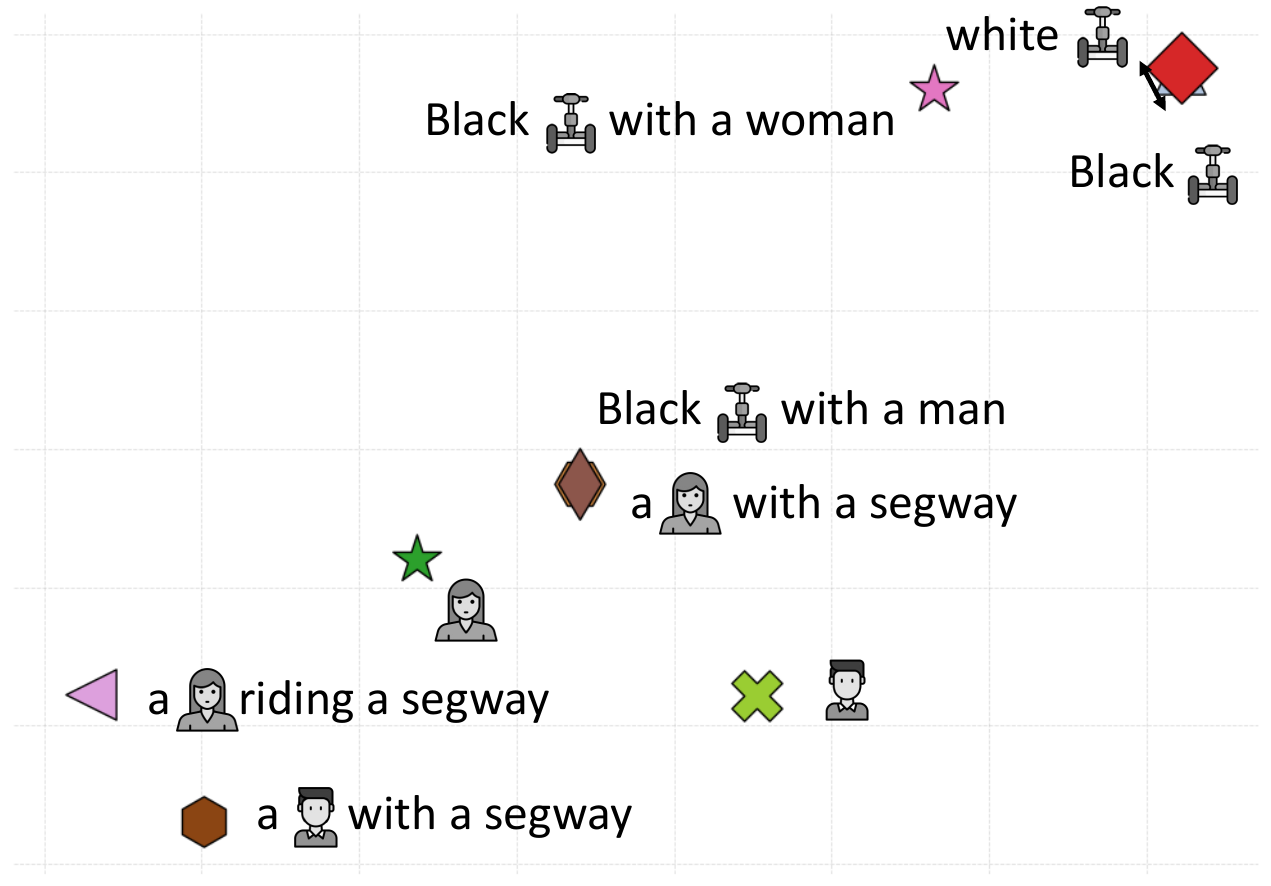}
\caption{E5-V text embedding}
\label{fig:ev5}
\end{subfigure}

\vspace{-2mm}
\caption{
Comparison of text embeddings across models and tiers.
(a) GLEE embedding visualization.
(b) \model embedding visualization.
(c) Zoomed view of \model embedding.
(d) SigLIP text embedding.
(e) E5-V text embedding.
}
\label{fig:embedding_comparison}

\vspace{-4mm}
\end{figure*}

\begin{figure*}[t]
\centering
\begin{subfigure}{0.32\linewidth}
    \centering
    \includegraphics[width=\linewidth]{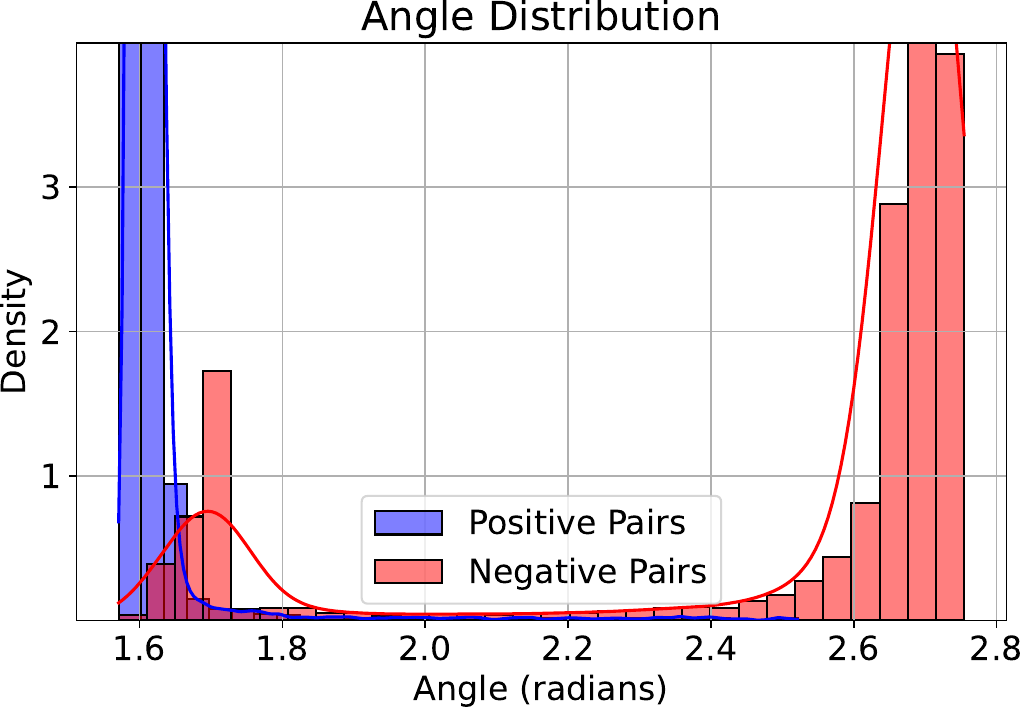}
    \caption{Tier 1}
\end{subfigure}
\hfill
\begin{subfigure}{0.32\linewidth}
    \centering
    \includegraphics[width=\linewidth]{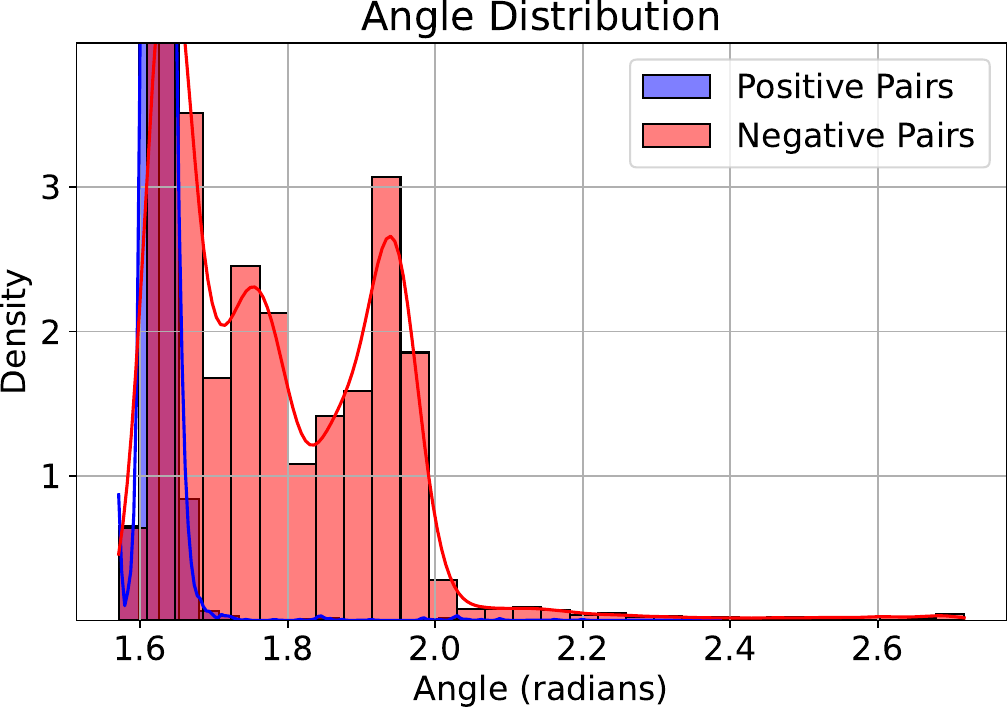}
    \caption{Tier 2}
\end{subfigure}
\hfill
\begin{subfigure}{0.32\linewidth}
    \centering
    \includegraphics[width=\linewidth]{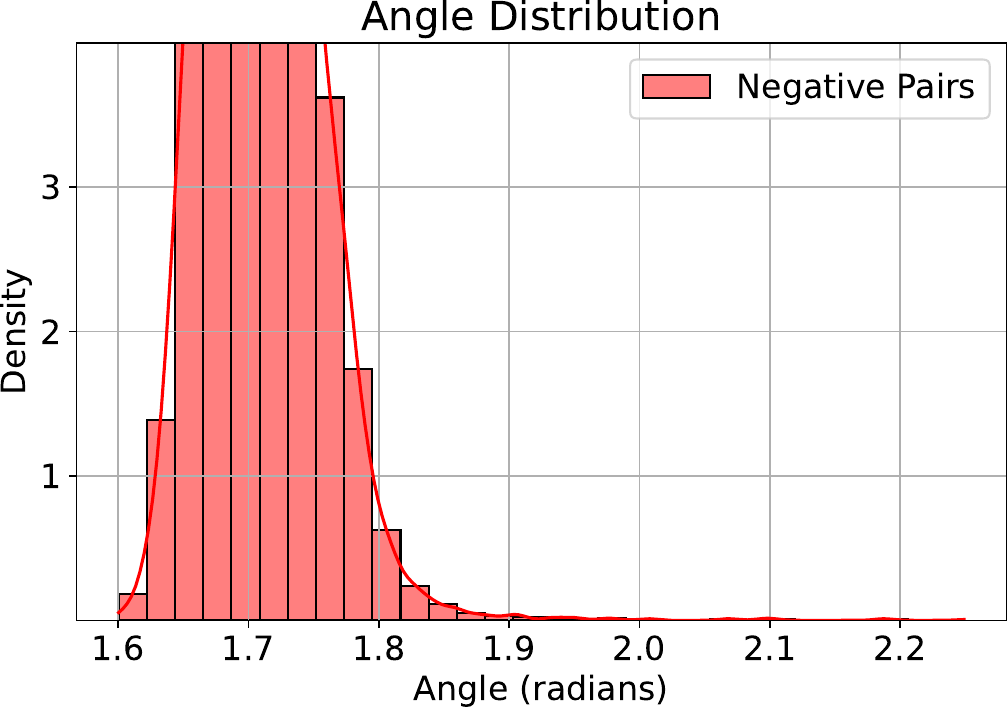}
    \caption{Tier 3}
\end{subfigure}
\caption{Exterior angle comparison across tiers. Each subplot shows the angular separation between positive and negative text embeddings.}
\label{fig:angle_tiers}
    \centering
    \includegraphics[width=0.85\linewidth]{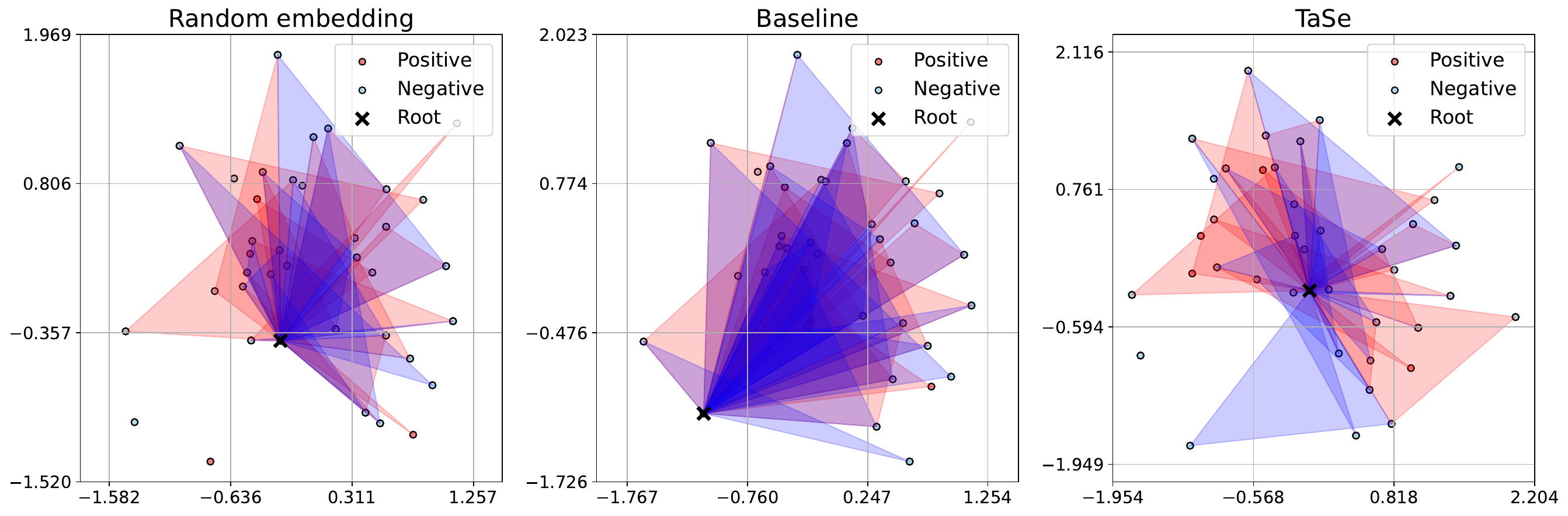}
    \caption{Visualization of angular embeddings (without dynamic reference). Triangles illustrate learned pairs with respect to the root: {\setlength{\fboxsep}{1pt}\colorbox{red!12}{positive pairs}} and {\setlength{\fboxsep}{1pt}\colorbox{blue!12}{negative pairs}} are connected to depict directional behavior. Positive pairs are expected to align in similar directions from the root, while negative pairs should diverge. While the baseline tends to increase radial distance more than meaningful angular adjustment, our objective function encourages more structured representations guided by directional alignment.}
    \label{fig:embedding}
\end{figure*}

We further analyze the exterior angles of text embeddings after training in Fig.~\ref{fig:angle} and Fig.~\ref{fig:angle_tiers}. 
Fig.~\ref{fig:angle} compares the embedding distributions of GLEE and TaSe. 
For GLEE, positive and negative exterior angles show substantial overlap, suggesting less distinct boundaries between positive and negative pairs. 
TaSe shows a more distinct separation of positive and negative pairs in angle space. 
To examine negative pairs with small angles, we compare angle distributions across semantic tiers in Fig.~\ref{fig:angle_tiers}.
Figures (a), (b), and (c) present Tier~1, Tier~2, and Tier~3 results. 
Each figure reports exterior angles between positive and negative samples.
Tier 1 shows a large angular separation between positive and negative samples. 
Tier 2 and Tier 3 show smaller separations for negative samples. 
Shared embeddings and pre-trained angular structures explain this pattern. 
The proposed method increases angular separation compared to the initial configuration. 
Positive and negative embeddings show meaningful disentanglement after training.

\subsection{Analysis of Hierarchical Objectives}
To assess the efficacy of the proposed $H^{+}$ and $H^{-}$ objectives in hierarchical representation learning, we performed experiments on synthetic datasets. Fifty pairs were randomly labeled as positive or negative with respect to a root, and the resulting embeddings were optimized using either the baseline or the proposed loss function with a fixed random seed. As shown in Fig.~\ref{fig:embedding}, random initialization fails to produce structured representations. The baseline objective causes pairs to drift away from both the root and each other. In contrast, the proposed objectives maintain proximity to the root while ensuring clear separation by clustering positive pairs and distancing negative ones.

%To verify whether disentangled embeddings contain distinct embedding representations for each component, we visualize the embedding of each component using t-SNE. For the t-SNE visualization, we construct the embedding space using our HiVG dataset of 10K samples. We then visualize the embeddings based on the language queries that motivated us. As shown in Fig.~\ref{fig:disentangled_tsne}, although there are slight variations across tokens, the embeddings for each component cluster relatively well. This validates that when a sentence is input, each component holds disentangled representations.

% 

\subsection{Failure Cases}
\begin{figure*}
    \centering
    \includegraphics[width=0.9\linewidth]{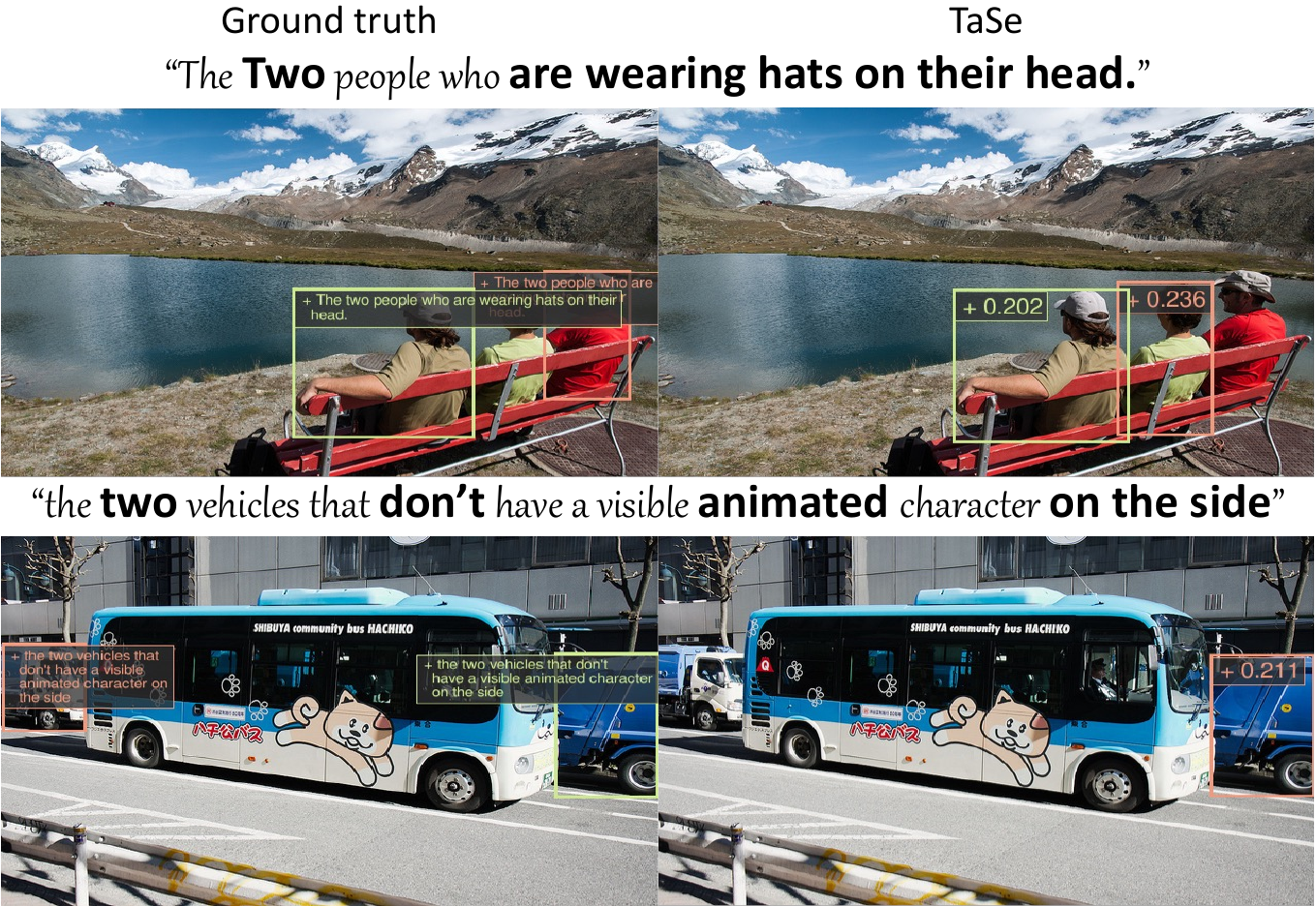}
    \includegraphics[width=0.9\linewidth]{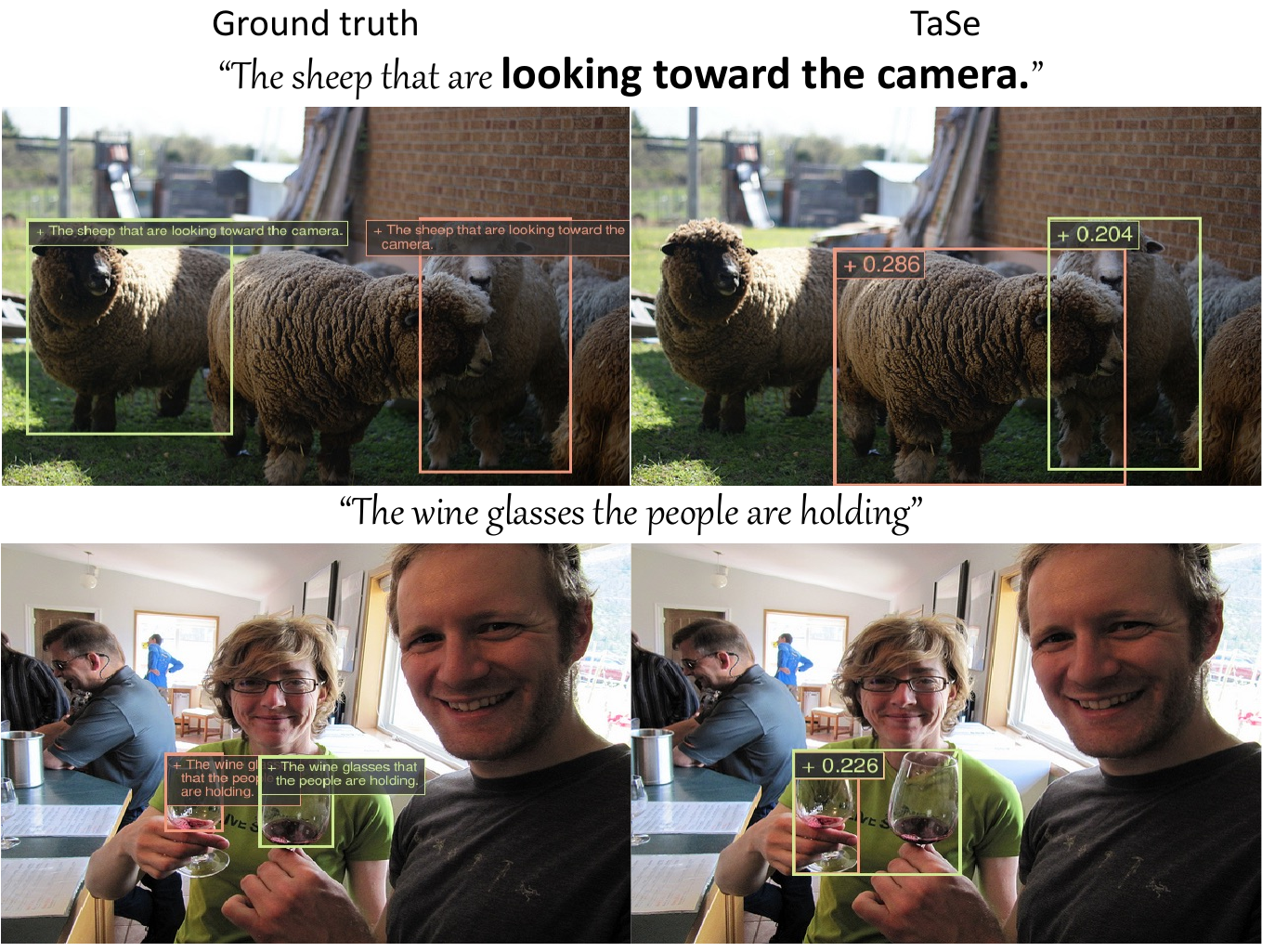}
    \caption{Qualitative results for failure cases.}
    \label{fig:failure}
\end{figure*}

We present qualitative results that indicate directions for improving the model.
As shown in Fig.~\ref{fig:failure}, accurately detecting objects remains challenging when additional attributes or relations are included.
Natural language contains complex structures, such as negation or compound structures, beyond the O-A-R components.
These diverse forms make it difficult to understand finer-grained semantic meaning, and this remains an important research problem.
TaSe reveals two major failure patterns: i) The model struggles with ambiguous expressions (e.g., looking toward the camera); and ii) Datasets such as RefCOCO contain only a single instance per object category, often resulting in predictions that merge multiple instances into one bounding box.


%\begin{wraptable}{r}{0.45\textwidth}
%\vspace{-10pt}
%\input{tabs/ablation}
%\input{tabs/disentangle}
%\vspace{-15pt}
%\end{wraptable}

%\begin{figure*}
%\centering
%\begin{subfigure}{0.7\linewidth}
%    \includegraphics[width=\linewidth]{figs/disentangle_tsne1.pdf}
%    \subcaption{Embedding visualization of the three disentangled components for the language query ``Segway with a man''}
%    \includegraphics[width=\linewidth]{figs/disentangle_tsne2.pdf}
%    \subcaption{Embedding visualization of the three disentangled components for the language query ``middle woman with dark hair''}
%\end{subfigure}
%\caption{t-SNE visualization of disentangled text embedding}
%\label{fig:disentangled_tsne}
%\end{figure*}

\end{document}